\documentclass{article}
\usepackage{microtype}
\usepackage{graphicx}
\usepackage{subcaption}
\usepackage{booktabs}
\usepackage{tabularx}
\usepackage{hyperref}
\usepackage{natbib}
\usepackage[ruled,vlined]{algorithm2e}
\usepackage{booktabs}
\usepackage{tabularx}
\usepackage{array} 
\usepackage[most]{tcolorbox}
\usepackage{minitoc}
\usepackage[preprint]{icml2026}

\usepackage{amsmath,amssymb,mathtools,amsthm,bm}
\usepackage[capitalize,noabbrev]{cleveref}
\usepackage{enumitem}

\usepackage[most]{tcolorbox}

\tcbset {
  base/.style={
    arc=0mm, 
    bottomtitle=0.5mm,
    boxrule=0mm,
    colbacktitle=black!10!white, 
    coltitle=black, 
    fonttitle=\bfseries, 
    left=2.5mm,
    leftrule=1mm,
    right=3.5mm,
    title={#1},
    toptitle=0.75mm, 
  }
}

\definecolor{brandblue}{rgb}{0.34, 0.7, 1}
\newtcolorbox{mainbox}[1]{
  colframe=brandblue, 
  base={#1}
}

\newtcolorbox{subbox}[1]{
  colframe=black!30!white,
  base={#1}
}

\theoremstyle{plain}

\theoremstyle{definition}

\theoremstyle{remark}

\newcommand{\E}{\mathbb{E}}
\newcommand{\R}{\mathbb{R}}

\newcommand{\defeq}{\triangleq}

\newcommand{\J}{J}

\icmltitlerunning{The Blessing of Dimensionality in LLM Finetuning: A Variance--Curvature Perspective}

\begin{document}

\twocolumn[
\icmltitle{The Blessing of Dimensionality in LLM Fine-tuning: \\A Variance--Curvature Perspective}

\begin{icmlauthorlist}
  \icmlauthor{Qiyao Liang}{mit}
  \icmlauthor{Jinyeop Song}{mit}
  \icmlauthor{Yizhou Liu}{mit}
  \icmlauthor{Jeff Gore}{mit}
  \icmlauthor{Ila Fiete}{mit}
  \icmlauthor{Risto Miikkulainen}{cognizant,ut}
  \icmlauthor{Xin Qiu}{cognizant}
\end{icmlauthorlist}

\icmlaffiliation{mit}{Massachusetts Institute of Technology, Cambridge MA, USA 02139}
\icmlaffiliation{cognizant}{Cognizant AI Lab, San Francisco CA, USA 94105}
\icmlaffiliation{ut}{The University of Texas at Austin, Austin TX, USA 78712}

\icmlcorrespondingauthor{Qiyao Liang}{qiyao@mit.edu}

\icmlkeywords{evolution strategies, fine-tuning, intrinsic dimension, Hessian spectrum, stochastic optimization, RLHF}

\vskip 0.3in
]

\printAffiliationsAndNotice{}  


\begin{abstract}
Weight-perturbation evolution strategies (ES) can fine-tune billion-parameter language models with surprisingly small populations (e.g., $N\!\approx\!30$), contradicting classical zeroth-order curse-of-dimensionality intuition. We also observe a second seemingly separate phenomenon: under fixed hyperparameters, the stochastic fine-tuning reward often rises, peaks, and then degrades in both ES and GRPO. We argue that both effects reflect a shared geometric property of fine-tuning landscapes: they are \emph{low-dimensional in curvature}. A small set of high-curvature dimensions dominates improvement, producing (i) heterogeneous time scales that yield rise--then--decay under fixed stochasticity, as captured by a minimal quadratic stochastic-ascent model, and (ii) degenerate improving updates, where many random perturbations share similar components along these directions. Using ES as a geometric probe on fine-tuning reward landscapes of GSM8K, ARC-C, and WinoGrande across Qwen2.5-Instruct models (0.5B--7B), we show that reward-improving perturbations remain empirically accessible with small populations across scales. Together, these results reconcile ES scalability with non-monotonic training dynamics and suggest that high-dimensional fine-tuning may admit a broader class of viable optimization methods than worst-case theory implies.
\end{abstract}

\begin{figure}[!htb]
  \centering
  \includegraphics[width=0.9\columnwidth]{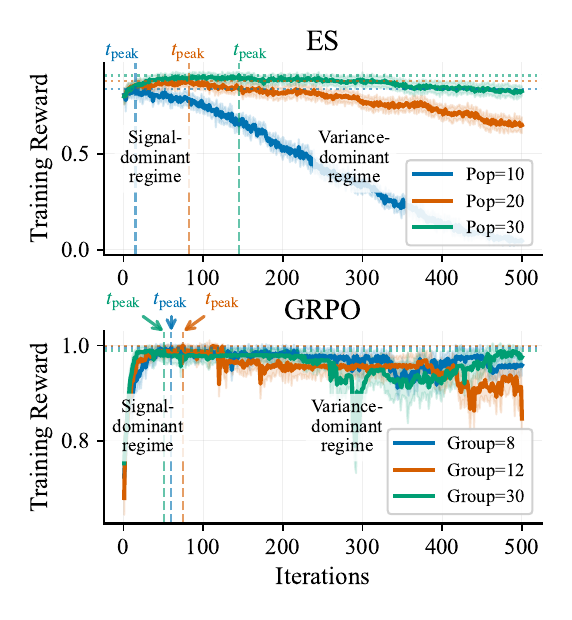}
  \caption{Non-monotonic training reward showing reward degradation beyond a peak training iteration $t_{\rm{peak}}$ in ES and GRPO fine-tuning of \texttt{Qwen2.5-1.5B-Instruct} on GSM8K~\cite{gsm8k}. Since GRPO does not directly maximize the task reward, training rewards plotted here are evaluated over the training set at every iteration. Both methods are trained on the same training set of 100 samples. }
  \label{fig:exp_nonmono}
\end{figure}

\section{Introduction}

Recent work has shown that large language models (LLMs) can be fine-tuned at scale using simple weight-perturbation evolution strategies~\citep[ES;][]{qiu2025es_at_scale}. This approach achieves performance competitive with leading reinforcement-learning-based methods such as Proximal Policy Optimization (PPO)~\cite{ppo} and Group Relative Policy Optimization (GRPO)~\cite{shao2024deepseekmath}. 
This is surprising because ES estimates an update direction from a finite set of random parameter perturbations; in generic high-dimensional problems, the signal-to-noise ratio of such estimates typically deteriorates with dimension, so avoiding a \emph{curse of dimensionality} would seem to require population sizes that grow rapidly with the number of parameters~\citep{duchi2015zero_order_rates}. 
Nonetheless, ES with populations as small as $N\approx 30$ has been shown to effectively fine-tune models with billions of parameters.

At the same time, we observe a second, seemingly independent phenomenon.
Under fixed hyperparameters over training duration, training reward often exhibits a pronounced non-monotonic trajectory: reward improves rapidly from a pretrained checkpoint, reaches a peak, and then \emph{decreases} toward a lower value as training proceeds.
This behavior appears consistently across ES and policy-gradient-style fine-tuning methods and is strongly modulated by algorithmic stochasticity—through the perturbation scale and population size in ES, or equivalently the temperature and group size in GRPO (Fig.~\ref{fig:exp_nonmono}).
Its ubiquity across distinct stochastic learning methods suggests that it reflects a structural property of the fine-tuning landscape itself, rather than an artifact of any particular algorithm.

In this paper, we aim to explain: (1) why can small-population weight-perturbation methods scale to billion-parameter models, and (2) why do stochastic fine-tuning runs exhibit rise--then--decay behavior under fixed stochasticity? To these ends, we posit the following hypothesis:

\begin{mainbox}{Central hypothesis: a blessing of dimensionality}

We hypothesize that LLM fine-tuning landscapes are \textbf{\emph{low-dimensional in curvature}}:
local optimization geometry is governed by a small set of curvature-active directions,
and the size of this curvature-active set does \textbf{\emph{not}} grow proportionally with the number of model parameters,
so small-population weight-perturbation methods can remain effective at billion-parameter scale despite zeroth-order curse-of-dimensionality intuitions.

\end{mainbox}

\paragraph{A unifying geometry--variance framework.}
We interpret both ES scalability and non-monotonic fine-tuning dynamics through a geometry--variance perspective intrinsic to the reward landscape.
Locally, curvature is highly anisotropic: a small number of high-curvature ``\emph{stiff}'' directions dominate reward improvement, while the vast majority of directions are weakly curved or effectively ``\emph{flat}.''
ES serves as a \emph{stochastic probe} of this structure, with explicit control over perturbation scale and sampling variance, allowing us to study how stochastic updates interact with anisotropic curvature.
Because improvement is governed by the stiff subspace, random-perturbation methods need only access this restricted set of directions, rather than estimate a full-dimensional gradient, enabling effective updates without populations that scale with parameter count.
At the same time, the stiff--flat separation induces heterogeneous time scales: rapid early progress occurs along stiff directions, while persistent stochastic drift along flat directions can dominate at later times, producing the observed rise--then--decay behavior under fixed stochasticity.

\paragraph{Contributions.}
We make three concrete contributions:
\begin{itemize}[noitemsep,topsep=0pt,parsep=1pt,partopsep=0pt,leftmargin=12pt]
  \item \textbf{A testable ``blessing of dimensionality'' prediction.}
  We formalize curvature-active dimensionality and derive scaling predictions linking model size, stochasticity, and the population needed to access improving perturbations.

  \item \textbf{A mechanism for rise--then--decay.}
  We give an analytically tractable model showing how stiff--flat time-scale separation under fixed noise yields non-monotonic reward trajectories.

  \item \textbf{ES-based probes across scale.}
  We empirically measure extreme-value improvement under perturbations across tasks and model sizes, showing that improving perturbations remain accessible with small populations from 0.5B to 7B parameters.
\end{itemize}

The rest of the paper is organized as follows. Section~\ref{sec:related_work} reviews zeroth-order optimization and prior evidence for low-dimensional fine-tuning and anisotropic curvature.
Section~\ref{sec:es} introduces ES as a geometric probe via Gaussian smoothing.
Section~\ref{sec:nonmonotonic} characterizes rise--then--decay in ES/GRPO and presents a minimal model linking heterogeneous curvature and fixed noise to non-monotonic dynamics.
Section~\ref{sec:empirical_scaling} tests the resulting scaling predictions on GSM8K~\cite{gsm8k}, ARC-C~\cite{arc_c}, and WinoGrande~\cite{winogrande}.
Section~\ref{sec:discussion} discusses implications for curse-of-dimensionality intuitions and fine-tuning methods.

\section{Background}
\label{sec:related_work}

This section reviews prior evidence motivating a \emph{blessing of dimensionality} view of fine-tuning: despite enormous parameter spaces, adaptation often depends on far fewer effective degrees of freedom. We briefly cover (i) zeroth-order curse-of-dimensionality intuition, (ii) empirical/architectural evidence for low-dimensional adaptation (intrinsic dimension, parameter-efficient updates), (iii) curvature structure in overparameterized networks (bulk near zero with a few dominant directions), and (iv) related work on stochastic dynamics and non-monotonic training. Together, these threads motivate our curvature-based notion of effective dimensionality and the use of ES as a stochastic probe of fine-tuning geometry.

\paragraph{Zeroth-order optimization and the curse of dimensionality.}
Classical results in zeroth-order (gradient-free) optimization often show that the number of function evaluations needed to make progress grows quickly with the number of parameters, which has led to skepticism about scaling weight-perturbation methods to modern overparameterized models~\citep{duchi2015zero_order_rates}.
Despite this, evolution strategies have long been studied as scalable black-box optimizers in reinforcement learning and large-scale learning, including distributed ES variants~\citep{salimans2017evolution}.
Recent work demonstrates that simple weight-perturbation ES can fine-tune billion-parameter language models with surprisingly small populations and competitive performance~\citep{qiu2025es_at_scale}.
Our work addresses the resulting tension between worst-case zeroth-order intuition and empirical scalability, and provides an operational lens that links scalability and training dynamics.

\paragraph{Blessing of dimensionality via low-dimensional fine-tuning structure.}
A growing body of evidence suggests that many fine-tuning problems are effectively low-dimensional despite extremely high ambient parameter dimension.
Intrinsic dimension studies show that downstream objectives can be optimized within a surprisingly small random subspace, including in language-model fine-tuning, and reporting that this intrinsic dimensionality actually decreases as a function of model size~\citep{li2018intrinsic_dimension_objective_landscapes,aghajanyan2021intrinsic_dim_lm_ft}.
Parameter-efficient fine-tuning methods provide architectural evidence for low-dimensional adaptation, for example via low-rank updates as in LoRA~\citep{hu2021lora}.
These results motivate a ``blessing of dimensionality'' viewpoint: additional parameters need not increase the number of directions that matter for adaptation, and may instead enlarge flat or redundant directions, resulting in a lower effective dimensionality. 

\paragraph{Hessian spectra: bulk near zero and a few outliers.}
A standard way to characterize local geometry in optimization is via the \emph{Hessian}, the matrix of second derivatives of the objective with respect to parameters.
Near a solution, the Hessian eigenvalues describe curvature along different directions: large-magnitude eigenvalues correspond to \emph{stiff} directions where the objective changes rapidly, while near-zero eigenvalues correspond to \emph{flat} directions.
Empirical studies of overparameterized neural networks have repeatedly found a highly structured Hessian spectrum near trained solutions, typically consisting of a large \emph{bulk} of eigenvalues concentrated near zero together with a small number of \emph{outliers} that dominate curvature~\citep{sagun2017hessian_overparam,ghorbani2019hessian_density}.
Related observations suggest that learning dynamics often concentrate in a low-dimensional ``sharp'' subspace associated with these top-curvature directions~\citep{gurari2018gradient_subspace}.
This bulk+outlier picture aligns with the geometric mechanism emphasized in our work---few stiff modes drive rapid early progress, while a large flat bulk is susceptible to variance accumulation---and motivates focusing on operational signatures of curvature structure rather than explicit Hessian estimation. 
Accordingly, we do not attempt to measure full Hessian spectra in our experiments; instead, we probe their practical consequences for stochastic fine-tuning, namely whether improvement depends on a small set of directions and how stochasticity couples to the flat bulk over time.

\paragraph{Stochastic optimization dynamics and noise-limited behavior.}
A complementary line of work models constant-step stochastic optimization near minima as a noisy dynamical system, often approximated by an Ornstein--Uhlenbeck process or related SDE, yielding predictions about stationary noise floors and curvature-dependent behavior~\citep{mandt2017sgd_as_bayes,li2017stochastic_modified_equations}.
These perspectives motivate interpreting late-stage training as variance-limited once gradient signal diminishes.
Our toy model adopts this analysis in the same spirit, using a local stochastic-dynamics approximation to make explicit how curvature and noise together shape late-stage behavior, and to explain how non-monotonic rise--then--decay can arise under fixed stochasticity.

\paragraph{Rise--then--decay and non-monotonic training dynamics.}
Non-monotonic training trajectories—where performance improves early and later degrades or oscillates under continued updates—have been documented in several stochastic learning settings.
In deep RL, extended on-policy training can exhibit performance collapse (often discussed as policy collapse or loss of plasticity), where returns improve initially and later deteriorate~\citep{dohare2023policycollapse,moalla2024notrust}.
In supervised deep learning, related non-monotonicity of the \emph{training loss} has been studied in ``edge of stability'' regimes, where loss can go up and down across iterations despite continued training~\citep{arora2022eos,zhu2022eos}.
In our setting, we observe a reproducible peak--then--decay phenomenon across both ES and policy-gradient-style fine-tuning (e.g., GRPO), suggesting a mechanism that is not specific to a particular optimization method.

\section{Evolution Strategies as a Geometric Probe}
\label{sec:es}

We use weight-perturbation evolution strategies (ES) primarily as a \emph{geometric probe} of fine-tuning landscapes.
Here a \emph{fine-tuning landscape} is the mapping from parameters to a scalar task reward,
$\theta \mapsto \mathcal{J}(\theta)$, where $\mathcal{J}$ may be accuracy or another reward evaluated via sampling and decoding.
ES is attractive in this setting because it requires only reward evaluations, making it applicable even when the underlying reward is discrete, truncated, or otherwise not amenable to differentiation.

At iteration $t$, ES samples perturbations $\varepsilon_k\sim\mathcal{N}(0,I)$, evaluates $r_k=\mathcal{J}(\theta_t+\sigma\varepsilon_k)$, and updates
\begin{equation}
  \widehat g_t \;=\; \frac{1}{N\sigma}\sum_{k=1}^N r_k\,\varepsilon_k,
  \qquad
  \theta_{t+1} \;=\; \theta_t + \alpha\,\widehat g_t,
  \label{eq:es_update_main}
\end{equation}
with the full procedure in Algorithm~\ref{alg:es}. We note that this is a naive version of the algorithm. For LLM fine-tuning in practice, we evaluate each candidate perturbation on a group of prompts and use the group-averaged reward in place of a single-sample evaluation.

\begin{algorithm}[t]
\caption{Weight-Perturbation Evolution Strategies (ES)}
\label{alg:es}
\KwIn{
  objective/reward $\mathcal{J}(\theta)$;
  initial parameters $\theta_0$;
  step size $\alpha$;
  perturbation scale $\sigma$;
  population size $N$;
  iterations $T$
}
\For{$t \gets 0$ \KwTo $T-1$}{
  Sample $\varepsilon_1,\ldots,\varepsilon_N \overset{\text{i.i.d.}}{\sim} \mathcal{N}(0,I)$\;

  \For{$k \gets 1$ \KwTo $N$}{
    $r_k \gets \mathcal{J}\!\left(\theta_t + \sigma \varepsilon_k\right)$\;
  }

  $\widehat{g}_t \gets \frac{1}{N\sigma}\sum_{k=1}^N r_k\,\varepsilon_k$, \;
  $\theta_{t+1} \gets \theta_t + \alpha\,\widehat{g}_t$\;
}
\end{algorithm}

\paragraph{Gaussian smoothing and coarse-grained geometry.}
ES can be interpreted as optimizing a Gaussian-smoothed objective,
\begin{equation}
  \mathcal{J}_{\sigma}(\theta)
  \;\defeq\;
  \mathbb{E}_{\varepsilon \sim \mathcal{N}(0,I)}\!\left[\mathcal{J}(\theta + \sigma \varepsilon)\right],
  \label{eq:gaussian_smoothing}
\end{equation}
where $\sigma$ controls the smoothing scale.
Even when $\mathcal{J}$ is jagged or nondifferentiable, $\mathcal{J}_{\sigma}$ is differentiable under mild conditions.
Geometrically, smoothing suppresses fine-scale irregularities while preserving curvature structure at scales larger than $\sigma$, so ES interacts with a coarse-grained version of the landscape.

A central identity is
\begin{equation}
  \nabla \mathcal{J}_{\sigma}(\theta)
  \;=\;
  \frac{1}{\sigma}\,
  \mathbb{E}_{\varepsilon \sim \mathcal{N}(0,I)}\!\left[\mathcal{J}(\theta+\sigma\varepsilon)\,\varepsilon\right],
  \label{eq:es_grad_identity}
\end{equation}
so $\widehat g_t$ is a Monte Carlo estimator of $\nabla \mathcal{J}_{\sigma}(\theta_t)$.

\paragraph{Controllable Stochasticity.}
With a finite population, ES produces a stochastic gradient estimate of $\nabla \J_\sigma$:
\begin{equation}
  \widehat g_t = \nabla \J_\sigma(\theta_t) + \xi^{\mathrm{est}}_t,
  \qquad \E[\xi^{\mathrm{est}}_t\mid\theta_t]=0,
  \label{eq:es_noise_decomp}
\end{equation}
where $\xi^{\mathrm{est}}_t$ is Monte Carlo estimation noise. Its covariance has the standard prefactor
\begin{equation}
  \mathrm{Cov}(\xi^{\mathrm{est}}_t\mid\theta_t)
  \;\approx\;
  \frac{1}{N\sigma^2}\,\Sigma_{\mathrm{est}}(\theta_t,\sigma),
  \label{eq:es_estimator_var}
\end{equation}
with problem-dependent $\Sigma_{\mathrm{est}}$ that can itself depend on $\sigma$ (e.g., through reward variability).
Separately, in our later local dynamics toy model we summarize the \emph{effective} stochasticity of parameter updates by a diffusion scale $\kappa=\sigma^2/N$, which captures how larger perturbation radii and smaller populations increase parameter-space wandering.

\paragraph{Why analyze fine-tuning landscapes with ES?}
Our goal is not to compare optimizers, but to understand the geometric structure of fine-tuning landscapes and how it interacts with stochastic learning dynamics.
A key challenge is that the \emph{true} reward landscape for LLM fine-tuning is often jagged or defined implicitly through sampling (e.g., discrete accuracy, externally judged rewards), so gradients are unavailable or only accessible through surrogate losses that may obscure the underlying reward geometry.
ES provides a natural entry point because it operates directly on reward evaluations under random perturbations and, crucially, induces controlled Gaussian smoothing.
This smoothing probes the landscape at a tunable spatial scale and couples cleanly to curvature structure, making ES well-suited for studying coarse-grained geometry and the role of variance. Subsequent sections use ES-based analyses to reveal geometric signatures of fine-tuning landscapes, and then relate these signatures to the behavior of other RL fine-tuning methods.

\section{Non-Monotonic Training Dynamics and Low-Dimensional Curvature}
\label{sec:nonmonotonic}

We begin with a simple but striking phenomenon: during fine-tuning on GSM8K, the training reward need not increase monotonically under fixed hyperparameters.
Instead, reward often improves rapidly, reaches a peak, and then degrades toward a lower value.
Figure~\ref{fig:exp_nonmono} shows this behavior for ES at multiple population sizes: both the peak time and the depth of the decay vary systematically with stochasticity, suggesting a variance-controlled effect rather than overfitting or evaluation noise.
We observe qualitatively similar, though noisier, rise--then--decay behavior in motivating GRPO runs as well, indicating that the phenomenon is not specific to a particular learning method.

\begin{figure}[t]
  \centering
  \includegraphics[width=\columnwidth]{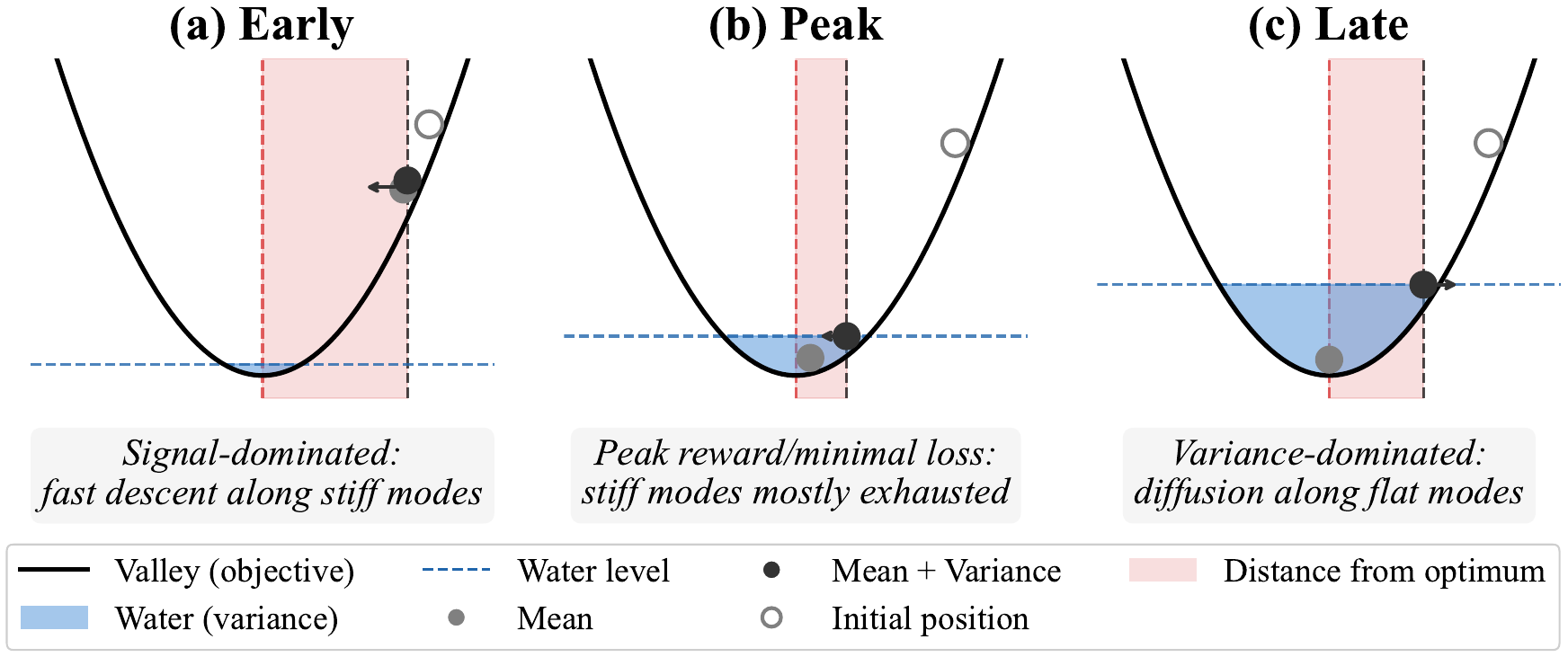}
  \caption{\textbf{Water-filling schematic for rise--then--decay dynamics.}
  \textbf{(a) Early:} fast improvement along stiff directions while variance is small.
  \textbf{(b) Peak:} stiff directions are mostly exhausted; variance has risen enough to limit gains.
  \textbf{(c) Late:} variance-dominated drift along weakly constrained directions yields degradation under fixed stochasticity.}
  \label{fig:valley_flooding}
\end{figure}

\paragraph{A mechanism isolate: local quadratic stochastic dynamics.}
To connect this behavior to landscape geometry, we analyze a local quadratic approximation around a near-optimal region.
Let $\theta^\star$ be a local maximizer and write
\begin{equation}
\J(\theta^\star + x) \approx \J(\theta^\star) - \tfrac{1}{2} x^\top C x,
\qquad C\succeq 0.
\end{equation}
A constant-step noisy ascent model takes the form
\begin{equation}
\theta_{t+1}=(I-\alpha C)\theta_t + \alpha\frac{\sigma}{\sqrt{N}}\varepsilon_t,
\qquad
\varepsilon_t\sim\mathcal N(0,I),
\label{eq:quad_update_main}
\end{equation}
with effective noise $\kappa=\sigma^2/N$.
Diagonalizing $C=Q\Lambda Q^\top$ decouples the dynamics into independent modes with contraction factors $a_i=1-\alpha\lambda_i$ (Appendix~\ref{app:toy_derivation}): high-curvature directions relax quickly, while low-curvature directions relax slowly and are more susceptible to variance accumulation.

\paragraph{Water-filling analogy: ``rushing downhill before the valley floods.''}
Figure~\ref{fig:valley_flooding} summarizes the core intuition in the equivalent \emph{descent} picture.
Early in training, fast relaxation along stiff directions drives rapid improvement.
Meanwhile, stochasticity accumulates along weakly constrained directions, which acts like a rising ``water level'' that progressively limits attainable performance.
Once the fast directions have largely saturated, the remaining dynamics can be dominated by this variance accumulation, producing a peak followed by degradation under fixed stochasticity.

\begin{figure}[t]
  \centering
  \includegraphics[width=0.9\columnwidth]{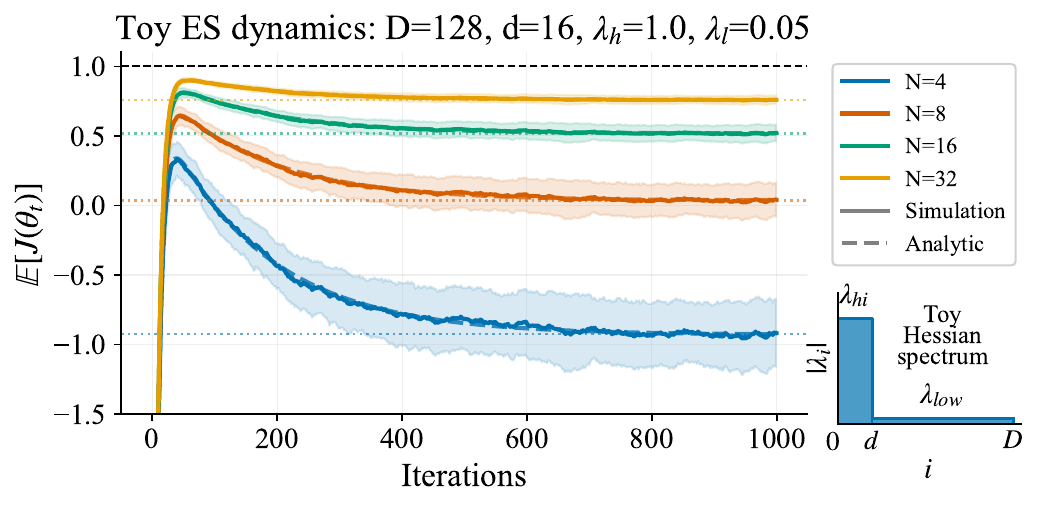}
  \caption{\textbf{Quadratic toy model: time-scale separation produces rise--then--decay.}
  ES on a quadratic landscape with a two-block spectrum ($D{=}128$, $d{=}16$, $\lambda_h{=}1.0$, $\lambda_l{=}0.05$).
  Solid curves: Monte Carlo ES runs for different populations $N$ (noise levels).
  Dashed curves: closed-form prediction from the quadratic stochastic model (Appendix~\ref{app:toy_derivation}).
  Larger $N$ (smaller $\kappa=\sigma^2/N$) raises the terminal plateau and suppresses late-time degradation.}
  \label{fig:toy_nonmono}
\end{figure}

\paragraph{A minimal example: two-block curvature spectrum.}
To make the time-scale separation concrete, we consider a stylized two-block spectrum: $d\ll D$ stiff directions with curvature $\lambda_{hi}$ and many weakly curved directions with $\lambda_{lo}\ll\lambda_{hi}$ (as shown in the inset of Figure~\ref{fig:toy_nonmono}). Figure~\ref{fig:toy_nonmono} shows ES dynamics (both simulation and analytics) on this toy quadratic landscape and demonstrates that simple spectral heterogeneity is sufficient to produce rise--then--decay trajectories, with early gains driven by the stiff subspace and late-time behavior set by noise accumulation in the weakly curved bulk. Moreover, in the quadratic case, variance tends to a terminal plateau value that is determined by the effective stochasticity dependent on $N$. 

\paragraph{What the toy model isolates.}
The toy model isolates two qualitative requirements for rise--then--decay in this local regime.
First, \emph{heterogeneous and low-dimensional curvature spectrum}: with a single dominant time scale the expected trajectory is monotonic, while well-separated time scales can produce a peak.
Second, \emph{non-negligible stochasticity}: the late-time plateau is controlled by $\kappa=\sigma^2/N$ through a curvature-weighted functional (Appendix~\ref{app:toy_derivation}).
Initialization affects how pronounced the peak is (Appendix~\ref{app:toy_derivation}), but the mechanism itself is simply the interaction of curvature-dependent relaxation and fixed noise. We note in passing that the plateau phenomenon naturally prescribes a curvature-based effective dimensionality measure, which inspires the proposal of a spectroscopy method that we describe in Appendix~\ref{app:clss_slq}.

\begin{figure}[t]
  \centering
  \includegraphics[width=0.9\columnwidth]{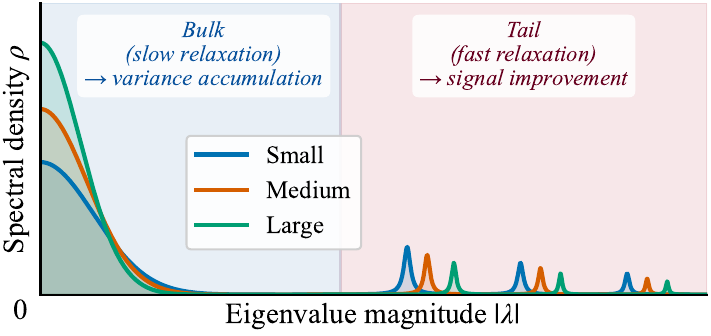}
  \caption{\textbf{Schematic curvature structure: near-zero bulk with a few stiff outliers.}
  The near-zero bulk corresponds to many weakly constrained directions, while a small number of outliers correspond to curvature-dominant directions that relax quickly and drive early improvement.}
  \label{fig:hessian_schematic}
\end{figure}

\paragraph{Implication: low-dimensional curvature in fine-tuning landscapes.}
The variance-controlled rise--then--decay behavior observed in ES (and qualitatively in our motivating GRPO runs) points to the same ingredients in real fine-tuning: a small number of fast, curvature-dominant directions and many weakly constrained directions.
This aligns with the common ``bulk + outliers'' curvature picture in overparameterized networks, where a few large-magnitude eigenvalues dominate curvature amid a near-zero bulk. An schematic illustration of the Hessian spectrum is given in Figure~\ref{fig:hessian_schematic}.
In our curvature-based notion of low-dimensionality, what matters is not the ambient parameter count but how many directions are meaningfully curvature-active, and that the curvature-active dimensions do not scale with model size.

\paragraph{From low-dimensional curvature to degeneracy.}
Nevertheless, it is important to note that low-dimensional curvature does not imply a unique improving direction.
Rather, if improvement is governed by a small curvature-active subspace, then many distinct perturbations can produce comparable progress by sharing similar components in that subspace.
In the next section, we show that this degeneracy has a concrete empirical consequence: random perturbations can reliably access reward-improving directions with a small, fixed population size even as model dimension increases.

\section{Empirical Scalability of ES as a Consequence of Low-Dimensional Curvature}
\label{sec:empirical_scaling}

The preceding section showed that rise--then--decay dynamics may arise from a low-dimensional \emph{curvature} structure of fine-tuning landscapes, leading to a concrete prediction: if curvature-active structure is low-dimensional and persists across scales, then improvement should remain accessible under random perturbations even as the ambient parameter dimension grows.
We now elaborate on and test this prediction across model sizes.

\paragraph{Definitions: accessibility and degeneracy of reward-improving perturbations.}
Let $C$ denote a local curvature operator (e.g., $C=-\nabla^2 \J(\theta^\star)\succeq 0$ near a local maximizer), and let $U\in\R^{D\times k}$ span a $k$-dimensional \emph{curvature-active} subspace (e.g., top-eigen directions of $C$).
For an isotropic perturbation $\varepsilon\sim\mathcal N(0,I)$, define its projection $z=U^\top\varepsilon\in\R^k$.
If improvement is primarily determined by $z$, then an improving region $\mathcal A\subset\R^k$ induces an improvement-supporting set in the full space,
\begin{equation}
\mathcal G \;\defeq\; \{\varepsilon\in\R^D:\ U^\top\varepsilon\in\mathcal A\}.
\end{equation}
When $k\ll D$, many distinct perturbations share similar curvature-active components: for any fixed $z\in\mathcal A$, the preimage $\{\varepsilon:U^\top\varepsilon=z\}$ is an affine subspace of dimension $D-k$.
We refer to this many-to-one structure as \emph{degeneracy}.
Whether ES can reliably find improving perturbations is instead governed by \emph{accessibility}, i.e., the probability mass of $\mathcal G$ under the sampling distribution,
\begin{equation}
p_{\mathrm{imp}}(\sigma)\;\defeq\;\Pr(\varepsilon\in\mathcal G)=\Pr(U^\top\varepsilon\in\mathcal A),
\end{equation}
which depends only on the geometry of $\mathcal A$ in the $k$-dimensional projected space.
Thus low-dimensional curvature can yield a nontrivial improvement-supporting mass even when $k\ll D$, providing a mechanism by which fixed-population random search can avoid curse-of-dimensionality behavior.

\begin{figure}[t]
  \centering
  \includegraphics[width=0.95\columnwidth]{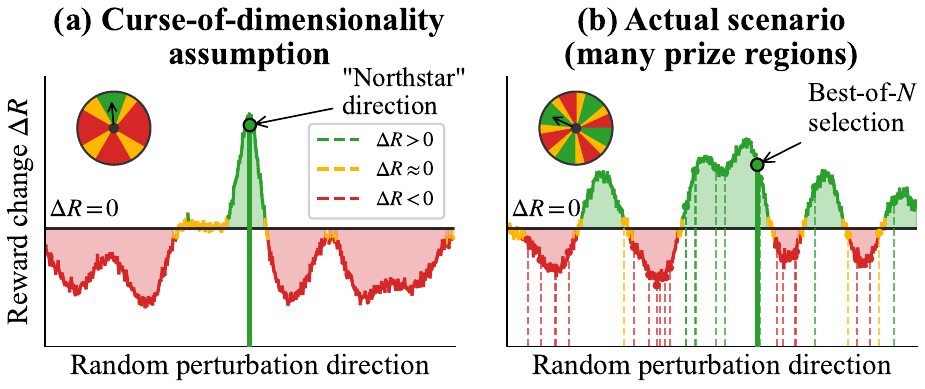}
  \caption{\textbf{Needle-in-a-haystack versus degenerate ``wheel of fortune.''}
  \textbf{(a)} In a classical curse-of-dimensionality intuition, improvement is confined to a single unique direction with vanishing probability mass, so fixed-population random search fails as dimension grows.
  \textbf{(b)} Under low-dimensional curvature, improvement is governed by a small curvature-active subspace, but many ambient perturbations share similar projections onto this subspace (degeneracy), yielding an improvement-supporting set with nontrivial probability mass.
  Extreme-value selection (best-of-$N$) can therefore succeed with a small, fixed population size.}
  \label{fig:cartoon_prize_regions}
\end{figure}
\paragraph{What an empirical curse of dimensionality would imply.}
Figure~\ref{fig:cartoon_prize_regions} contrasts two geometries.
In a needle-in-a-haystack picture (a), improvement lies in a vanishingly rare ``north-star'' direction, so a fixed population is unlikely to sample $\Delta R>0$ as dimension grows.
In a degenerate many--prize-region picture (b), improvement is not unique: many perturbation directions yield positive $\Delta R$, so best-of-$N$ selection can reliably find an improving update without requiring $N$ to grow with model size.
Empirically, a curse of dimensionality would therefore appear as loss of improvement signal at fixed $N$, or a systematic rightward shift of best-of-$N$ curves with scale; our experiments test for these signatures.

\paragraph{Metrics: best-of-$N$ as an operational proxy for accessibility.}
For each model and task, we sample perturbations $\varepsilon_i\sim\mathcal N(0,I)$ and measure
$\Delta R_i = R(\theta+\sigma\varepsilon_i)-R(\theta)$.
We summarize accessibility by the expected best-of-$N$ improvement
\begin{equation}
\Delta^*_N(\theta,\sigma)=\mathbb{E}\!\left[\max_{i\le N}\Delta R_i\right],
\end{equation}
which directly quantifies whether $N$ random draws can reach the improving ``prize regions'' in Fig.~\ref{fig:cartoon_prize_regions}(b), even when $\mathbb{E}[\Delta R]$ is small or negative.
To control for task saturation, we also report headroom-normalized improvements $\Delta^*_N/(1-R_0)$.

\begin{figure*}[t]
\centering
\includegraphics[width=0.68\textwidth]{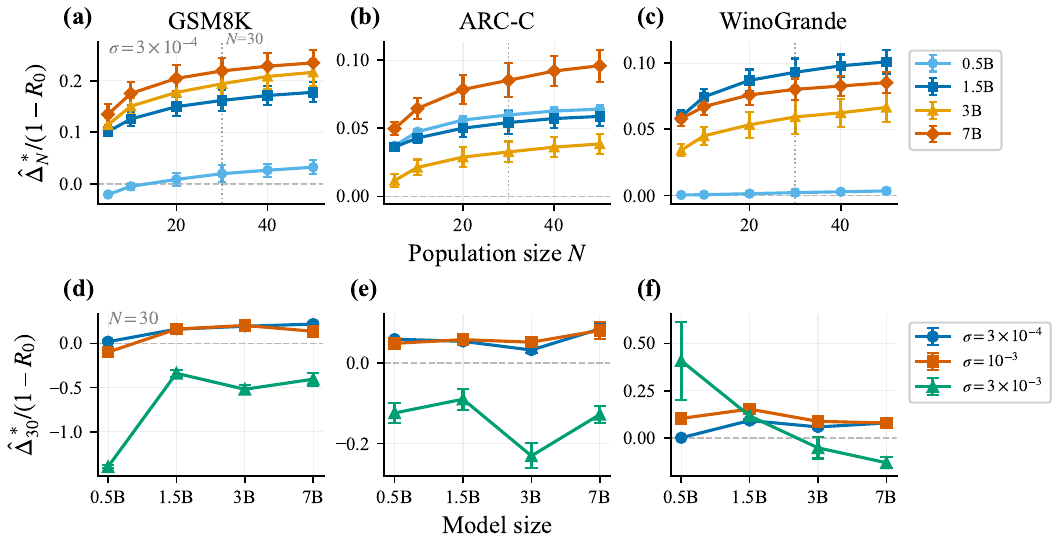}
\caption{\textbf{Fixed-population improvement remains accessible across model scales.}
\textbf{Top row (a--c):} Headroom-normalized expected best-of-$N$ improvement
$\Delta^{\mathrm{rel}}_N(\sigma)=\E[\max_{i\le N}\Delta R_i]/(1-R_0)$ as a function of population size $N$
at fixed perturbation scale $\sigma=3\times10^{-4}$ for GSM8K, ARC-C, and WinoGrande across \texttt{Qwen2.5-Instruct} (0.5B--7B).
Curves saturate by $N\approx 30$--40 without a systematic shift to larger $N$ as model size increases.
\textbf{Bottom row (d--f):} Headroom-normalized expected best-of-30 improvement versus model size for multiple $\sigma$.
For each task, a viable range of $\sigma$ yields positive best-of-30 improvements from 0.5B to 7B, indicating that the improvement-supporting tail of the perturbation-induced $\Delta R$ distribution remains accessible at scale.
\textbf{Error bars} show $\pm 1.96\,\mathrm{SE}$ computed across $S$ independent perturbation batches (each batch contains a fixed pool of candidates evaluated on the same prompt set), reflecting variability due to perturbation sampling (Appendix~\ref{app:exp_details_es_probe}).}
\label{fig:es_scaling_combined}
\end{figure*}

\paragraph{Population requirements do not scale with model size.}
Figures~\ref{fig:es_scaling_combined}(a--c) report the \emph{headroom-normalized expected best-of-$N$ improvement}
as a function of population size $N$ for GSM8K, ARC-C, and WinoGrande across the
\texttt{Qwen2.5-Instruct} family (0.5B--7B).
For each task, model, and perturbation scale $\sigma$, we construct a pool of $M=240$
perturbation candidates and evaluate their reward changes
$\Delta R = R(\theta+\sigma\epsilon)-R(\theta)$ on a fixed prompt set.
We estimate $\Delta_N^*(\sigma)$ by Monte Carlo sampling $N$ \emph{distinct} candidates
(without replacement) from this pool and averaging the resulting maxima, and report
the headroom-normalized quantity $\Delta_N^*(\sigma)/(1-R_0)$
(see Appendix~\ref{app:exp_details_es_probe} for experimental details).

Across all tasks and model sizes, best-of-$N$ improves rapidly at small $N$ and exhibits
clear diminishing returns beyond $N \approx 30$--40, with no systematic rightward shift
as model size increases.
This rules out the most direct empirical signature of a curse of dimensionality---namely,
a growing population requirement to access improvement as ambient parameter dimension
increases---and indicates that improvement remains accessible with relatively small
populations even for the largest models studied. Importantly, this saturation should not be attributed to a finite-candidate-pool artifact.
With $M=240$ candidates and $N \le 50$, we are far from the $N \to M$ regime in which
finite-pool saturation dominates.
Instead, the observed flattening reflects intrinsic diminishing returns of expected
extrema, which arise generically---even in the infinite-pool (i.i.d.) limit---from the
shape of the improvement distribution’s upper tail.
In this regime, increasing population size primarily yields rarer tail events rather
than systematically larger gains.

\paragraph{Viable (local) perturbation scales persist with model size.}
Fixing the population size to $N=30$, Figures~\ref{fig:es_scaling_combined}(d--f) show the
headroom-normalized best-of-$30$ improvement $\Delta_{30}^*(\sigma)/(1-R_0)$ as a function
of model size for multiple perturbation scales $\sigma$.
Across all tasks, there exists a \emph{viable} range of sufficiently small perturbation
scales for which $\Delta_{30}^*(\sigma)$ remains positive from 0.5B to 7B parameters. The key requirement is not precise tuning of $\sigma$, but that the perturbations remain
local enough to probe regions of parameter space where improvement is common.
At these scales, the perturbation distribution intersects many improvement-supporting
regions, so that a moderate population reliably samples candidates from the upper tail.
While the magnitude of improvement and the optimal $\sigma$ vary by task (and absolute
gains necessarily attenuate as headroom shrinks), improvement does not collapse as the
number of parameters increases by more than an order of magnitude. Operationally, this indicates that a constant ES population budget ($N \approx 30$),
combined with sufficiently local perturbations, continues to access an
improvement-supporting tail of the landscape at scale.

\paragraph{Interpretation: a variance--geometry tradeoff yields a blessing of dimensionality.}
Taken together, the preceding results support an interpretation in which high-dimensional
parameter spaces are \emph{not} hostile to zeroth-order search, but instead contain many
distinct, locally improving directions that can be accessed with modest populations.
At a checkpoint $\theta$ and perturbation scale $\sigma$, the local geometry of the reward
surface induces a distribution of perturbation outcomes $\Delta R$, whose upper tail
encodes the density of improvement-supporting regions.
ES progress at population $N$ is therefore governed by this \emph{tail mass}, summarized
by $\Delta_N^*(\sigma)$, rather than by alignment with a single gradient direction. Appendix~\ref{app:blessing_diag} reports auxiliary tail-accessibility summaries ($N_{90}$ and $q_{0.95}$) that likewise show no systematic growth with model size. 

Crucially, the persistence of best-of-$N$ improvement with fixed $N$ and suitably small
$\sigma$ across model sizes indicates that this improvement-supporting tail does not thin
as ambient dimension increases.
Instead, increasing dimensionality appears to introduce \emph{more} locally improving
directions, yielding a form of blessing of dimensionality in which improvement remains
accessible through random sampling. 
Finite populations succeed not by resolving a unique descent direction, but by reliably
intersecting one of many favorable regions. We note that although we do not go to the extreme to claim that larger models are lower-dimensional, complementary ES-based curvature proxy measurements suggest that curvature-relevant structure can become more concentrated with model scale (Appendix~\ref{app:reward_curvature_slq}).

\section{Discussions and Future Work}
\label{sec:discussion}

\paragraph{Central thesis: low-dimensional, heterogeneous, and degenerate fine-tuning geometry.}
Our results support a single geometric picture that reconciles two seemingly disparate empirical facts: (i) stochastic fine-tuning can exhibit rise--then--decay training dynamics under fixed hyperparameters, and (ii) small-population weight perturbation can improve billion-parameter LLMs without an empirical curse of dimensionality.
The unifying lens is that fine-tuning landscapes are effectively \emph{low-dimensional in curvature} yet \emph{heterogeneous} and \emph{degenerate}.
Low-dimensionality means that a small number of curvature-active (stiff) dimensions dominate progress; heterogeneity means that these directions have widely separated relaxation rates, producing fast early gains and slow variance accumulation; degeneracy means that improvement is not confined to a single unique direction but to a low-dimensional subspace that admits many equivalent embeddings in the ambient parameter space, yielding multiple improvement-supporting regions.
Together, these properties explain both non-monotonic training and ES scalability.

\paragraph{Practical consequence: diagnosing and mitigating rise--then--decay.}
The toy mechanism implies that rise--then--decay is not mysterious: it is the expected outcome when stiff modes saturate while variance continues to accumulate along flat modes under fixed stochasticity.
This suggests immediate practical interventions:
(i) \textbf{early stopping} (stop near the peak when stiff-mode signal is exhausted but variance has not yet dominated),
(ii) \textbf{noise scheduling} (increase population size, reduce $\sigma$, reduce temperature, or reduce effective update noise over time to slow ``water filling'' of flat modes),
and (iii) \textbf{adaptive step sizes} that shrink once curvature-active progress saturates.
More broadly, the geometry--variance view suggests treating non-monotonicity as a \emph{diagnostic}: its presence indicates heterogeneous curvature and a variance-dominated late regime, while its absence (under comparable stochasticity) suggests either insufficient heterogeneity or a regime in which variance is already controlled.

\paragraph{A broader implication: revisiting algorithm design beyond the curse-of-dimensionality dogma.}
Perhaps the most important consequence of our findings is conceptual.
Classical zeroth-order theory discourages parameter-space perturbation methods at high dimension by focusing on worst-case settings in which improvement is a single needle-like direction.
Our evidence instead supports a regime in which improvement directions are low-dimensional and degenerate, so a small population can reliably access them.
This opens a broader algorithmic design space that is typically dismissed for LLM fine-tuning:
\textbf{population-based perturbation methods}, \textbf{random subspace or coordinate search}, \textbf{structured perturbations} (e.g., low-rank or blockwise noise), \textbf{evolution strategies with learned search covariances}, and \textbf{hybrid methods} that combine occasional perturbative exploration with gradient updates.
In other words, once the effective geometry is low-dimensional, the relevant question is no longer ``can zeroth-order scale in $D$?'' but ``which perturbation distributions best align with the small set of curvature-active directions?''

\paragraph{Implications for robustness and diversity of solutions.}
The ``many prize regions'' interpretation suggests that different improving perturbations may lead to distinct solution manifolds with different properties (e.g., robustness, calibration, reasoning style, or safety tradeoffs), even when they yield comparable short-horizon reward gains.
This raises an opportunity: rather than searching for a single optimum, fine-tuning can be viewed as selecting among multiple alternative improvements.
Population-based methods are naturally suited to this perspective because they generate sets of candidate updates that can be filtered or diversified according to secondary objectives (such as conciseness, self-consistency, etc.).
Exploring this connection---how degeneracy relates to solution diversity and downstream robustness---is a promising direction for future work.

\paragraph{Limitations and scope.}
Our analysis is intentionally local and mechanism-driven.
The quadratic model isolates a \emph{sufficient} within-basin mechanism for rise--then--decay, but real fine-tuning is nonstationary and may traverse regions with changing curvature and noise structure; in particular, additional nonconvex effects such as saddle crossings or basin-to-basin drift can also contribute (Appendix~\ref{app:anisotropic_lyapunov}, \ref{app:beyond_quadratic}, and the double-well metastability discussion in Appendix~\ref{app:double_well}).
Our empirical probes use finite evaluation pools and task-specific rewards; while we quantify uncertainty and use headroom-normalized metrics, different evaluation protocols can shift absolute effect sizes.

\paragraph{Conclusion and future work.}
Our results point to a landscape-centric view of fine-tuning: the effective geometry that governs improvement can be low-dimensional in curvature and need not grow proportionally with model size. Looking ahead, three directions follow naturally. 
First, develop practical and scalable estimates of curvature-active structure---and test how it changes with model scale---in settings where objectives are reward-defined and gradients may be unreliable. 
Second, use these estimates to design perturbation distributions and population-based optimizers that better target the curvature-active subspace (e.g., structured or learned perturbations, subspace methods, and principled hybrids with gradient updates). 
Third, extend the variance--geometry framework to anisotropic and state-dependent noise to more directly connect ES-style probes with policy-gradient fine-tuning dynamics.



\bibliography{refs}
\bibliographystyle{icml2026}

\newpage
\appendix
\onecolumn

\part*{Appendix}
\addcontentsline{toc}{section}{Appendix}

\vspace{-0.5em}
\begin{tcolorbox}[
  enhanced,
  colback=gray!3,
  colframe=gray!45!black,
  boxrule=0.6pt,
  arc=1.5mm,
  left=1.2mm,right=1.2mm,top=0.9mm,bottom=0.9mm
]
\noindent\textbf{Appendix Table of Contents}

\vspace{0.35em}
\setlength{\tabcolsep}{4pt}
\renewcommand{\arraystretch}{1.12}
\begin{tabularx}{\linewidth}{@{}l X r@{}}
\textbf{A.} &
\hyperref[app:toy_derivation]{Quadratic OU Toy Model and Variance--Curvature Spectroscopy}\dotfill &
\pageref{app:toy_derivation} \\

\textbf{B.} &
\hyperref[app:clss_slq]{From Toy Slope Laws to Practical Curvature Probes: CLSS vs.\ SLQ}\dotfill &
\pageref{app:clss_slq} \\

\textbf{C.} &
\hyperref[app:anisotropic_lyapunov]{Anisotropic Noise and the Lyapunov Equation}\dotfill &
\pageref{app:anisotropic_lyapunov} \\

\textbf{D.} &
\hyperref[app:beyond_quadratic]{Beyond Quadratic: Saddles and Negative Curvature}\dotfill &
\pageref{app:beyond_quadratic} \\

\textbf{E.} &
\hyperref[app:double_well]{Metastability in a Double-Well: Escape Times and Hopping Criteria}\dotfill &
\pageref{app:double_well} \\

\textbf{F.} &
\hyperref[sec:appendix_countdown]{Additional Rise--Then--Decay Results: Countdown}\dotfill &
\pageref{sec:appendix_countdown} \\

\textbf{G.} &
\hyperref[app:exp_details_es_probe]{Experimental Details: ES Reward-Probe}\dotfill &
\pageref{app:exp_details_es_probe} \\
\end{tabularx}
\end{tcolorbox}

\section{Quadratic OU Toy Model and Variance--Curvature Spectroscopy}
\label{app:toy_derivation}

This appendix provides analytic details underlying the toy-model mechanism used in Section~\ref{sec:nonmonotonic}.
The main text uses this quadratic OU analysis as a \emph{mechanism isolate}: it shows that rise--then--decay dynamics arise generically from (i) heterogeneous curvature scales and (ii) fixed stochasticity, without requiring nonconvex pathologies.
We also show how the same closed-form expressions yield a ``spectroscopy'' viewpoint: the terminal plateau varies predictably with the effective noise scale $\kappa=\sigma^2/N$, and the slope encodes a curvature-weighted effective dimension.
This slope-fitting idea motivates the CLSS procedure introduced in the next appendix section (Section~\ref{app:clss_slq}).

\subsection{From noisy updates to a local quadratic OU model}
\label{app:toy_ou}

\paragraph{A generic noisy update abstraction.}
We consider maximizing an objective $\J(\theta)$ with a constant-step noisy update
\begin{equation}
  \theta_{t+1} = \theta_t + \alpha\bigl(g(\theta_t) + \xi_t\bigr),
  \label{eq:generic_update}
\end{equation}
where $\mathbb{E}[\xi_t\mid\theta_t]=0$ and $\mathrm{Cov}(\xi_t\mid\theta_t)=\Sigma(\theta_t)$.
For ES with isotropic perturbations, a useful local approximation is $\Sigma(\theta_t)\approx (\sigma^2/N) I$ (up to problem-dependent scaling), making the effective noise level explicitly tunable by $\kappa=\sigma^2/N$.
Policy-gradient methods (e.g., GRPO) have anisotropic $\Sigma(\theta_t)$; the analysis below extends by replacing $(\sigma^2/N)I$ with a general covariance and solving a Lyapunov equation, but we focus here on the isotropic case to isolate curvature effects.

\paragraph{Why a quadratic model?}
The rise--then--decay signature in Section~\ref{sec:nonmonotonic} indicates a regime where (i) gradients are small and (ii) stochasticity dominates long-time behavior.
In this regime, a local Taylor expansion around a critical point $\theta^\star$ is the natural first approximation:
\begin{equation}
\J(\theta^\star+x)\approx \J(\theta^\star)+\tfrac{1}{2}x^\top Hx,
\end{equation}
where $H=\nabla^2\J(\theta^\star)$.
Near a local maximizer, $H$ is negative semidefinite; we reparameterize curvature by $C\defeq -H\succeq 0$.
The quadratic model is not intended as a global description of LLM fine-tuning; rather, it yields exact mode-wise relaxation rates and noise floors, making the origin of non-monotonicity transparent.

\subsection{Quadratic toy model: exact dynamics and a criterion for non-monotonicity}
\label{app:toy_exact}

\paragraph{Setup.}
Let $\theta_t\in\mathbb{R}^D$ and define the quadratic reward
\begin{equation}
  \J(\theta)=1-\tfrac{1}{2}\theta^\top C\theta,
  \label{eq:J_quad}
\end{equation}
with maximizer $\theta^\star=0$.
Then $\nabla\J(\theta)=-C\theta$.
We model ES-like noisy ascent by
\begin{equation}
  \theta_{t+1} = (I-\alpha C)\theta_t + \alpha\frac{\sigma}{\sqrt{N}}\,\varepsilon_t,
  \qquad \varepsilon_t\sim\mathcal N(0,I_D),
  \label{eq:theta_update}
\end{equation}
and assume $\alpha\in(0,2/\lambda_{\max}(C))$ for stability on the range of $C$.

\paragraph{Diagonalization and mode-wise OU/AR(1).}
Let $C=Q\Lambda Q^\top$ with $\Lambda=\mathrm{diag}(\lambda_1,\dots,\lambda_D)$.
In the eigenbasis $x_t=Q^\top\theta_t$, modes decouple:
\begin{equation}
  x_{i,t+1} = a_i x_{i,t} + b\,\xi_{i,t},
  \qquad a_i=1-\alpha\lambda_i,\quad b=\alpha\sigma/\sqrt{N},\quad \xi_{i,t}\sim\mathcal N(0,1).
  \label{eq:ar1_update}
\end{equation}
Thus each spectral mode is an independent AR(1)/OU process.
For small $\alpha\lambda_i$, the characteristic decay time scales as $\tau_i\approx (\alpha\lambda_i)^{-1}$: stiff modes ($\lambda_i$ large) relax quickly while flat modes ($\lambda_i$ small) relax slowly.

\paragraph{Mean/variance dynamics and stationary noise floor.}
Let $\mu_{i,t}=\mathbb{E}[x_{i,t}]$ and $v_{i,t}=\mathrm{Var}(x_{i,t})$.
From~\eqref{eq:ar1_update}:
\begin{equation}
\mu_{i,t}=a_i^t\mu_{i,0},\qquad
v_{i,t}=a_i^{2t}v_{i,0} + \frac{b^2}{1-a_i^2}\bigl(1-a_i^{2t}\bigr).
\end{equation}
For $\lambda_i>0$ and $|a_i|<1$, the stationary variance is
\begin{equation}
  v_{i,\infty}=\frac{b^2}{1-a_i^2}
  =\frac{\alpha\sigma^2}{N\,\lambda_i(2-\alpha\lambda_i)}.
  \label{eq:vinf}
\end{equation}
The expected reward is
\begin{equation}
\mathbb{E}[\J(\theta_t)] = 1-\tfrac{1}{2}\sum_{i:\lambda_i>0}\lambda_i\bigl(\mu_{i,t}^2+v_{i,t}\bigr).
\end{equation}
At stationarity ($t\to\infty$), $\mu_{i,t}\to 0$ and $v_{i,t}\to v_{i,\infty}$, giving
\begin{equation}
  1-\J_\infty
  = \tfrac{1}{2}\sum_{\lambda_i>0}\lambda_i v_{i,\infty}
  = \frac{\alpha\sigma^2}{2N}\sum_{\lambda_i>0}\frac{1}{2-\alpha\lambda_i}.
  \label{eq:Jinf}
\end{equation}
Thus the terminal plateau is controlled by the effective noise $\kappa=\sigma^2/N$ and by a curvature-weighted trace functional of the spectrum.

\paragraph{Mixture-of-exponentials and non-monotonicity.}
Define amplitudes
\begin{equation}
  A_i \defeq -\tfrac{1}{2}\lambda_i\bigl(x_{i,0}^2-v_{i,\infty}\bigr), 
  \label{eq:Ai_def}
\end{equation}
assuming deterministic initialization $x_{i,0}$. 

Then the deviation from plateau is a mixture of exponentials:
\begin{equation}
  \mathbb{E}[\J(\theta_t)]-\J_\infty
  =\sum_{\lambda_i>0} A_i a_i^{2t}
  =\sum_{\lambda_i>0} A_i e^{-\gamma_i t},
  \qquad \gamma_i\defeq -2\log(|1-\alpha\lambda_i|).
  \label{eq:mixture_exp}
\end{equation}
Non-monotonicity occurs when the $\{A_i\}$ have mixed signs:
fast stiff modes can initially increase reward, while slow flat modes contribute delayed degradation as variance accumulates.
A pronounced peak arises when stiff modes begin far from equilibrium ($x_{i,0}^2\gg v_{i,\infty}$) while flat modes begin near or beyond their noise-limited equilibrium ($x_{i,0}^2\lesssim v_{i,\infty}$), matching the main-text initialization discussion.

\begin{figure*}[t]
  \centering
  \includegraphics[width=0.8\textwidth]{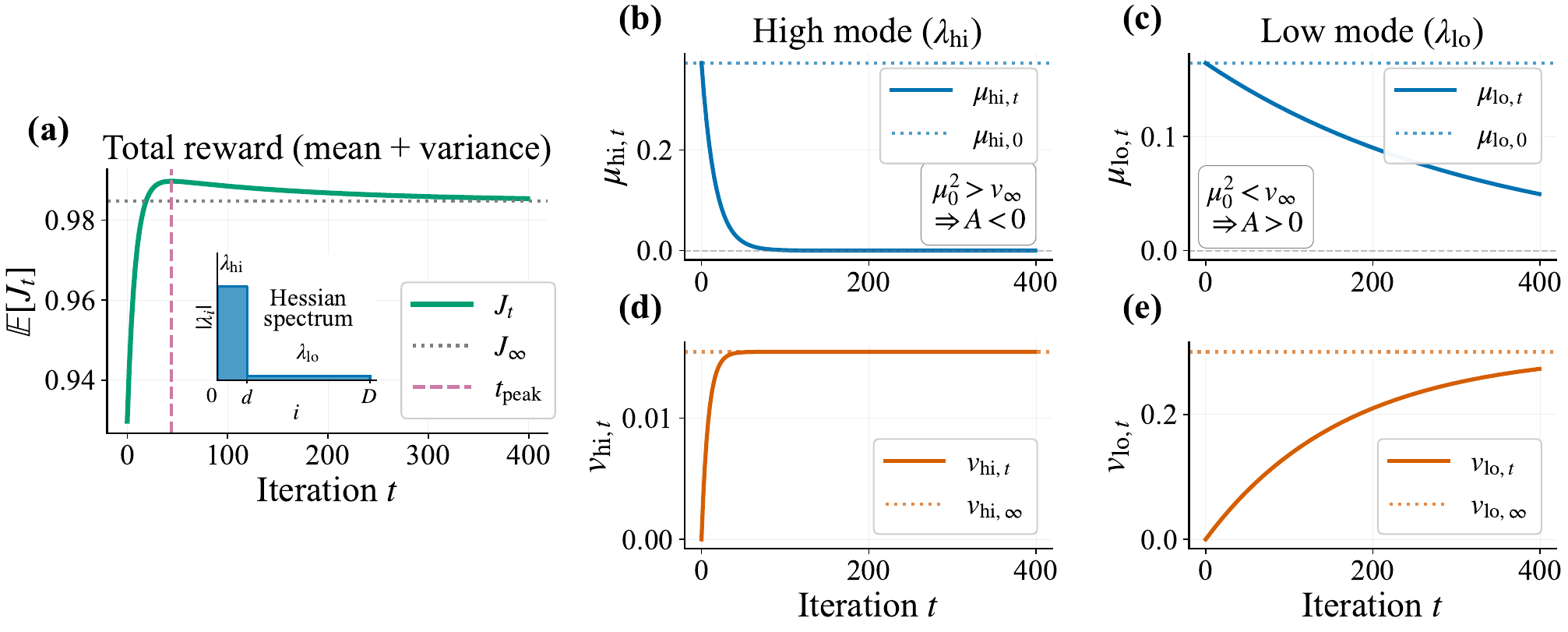}
  \caption{\textbf{Mode-wise relaxation rates explain rise--then--decay.}
  A two-mode illustration of~\eqref{eq:ar1_update}--\eqref{eq:mixture_exp} showing how curvature sets relaxation times.
  \textbf{(a)} Total reward $\mathbb{E}[J_t]$ peaks and then decays toward $J_\infty$ as slow variance continues to accumulate after stiff-mode signal is exhausted (vertical line marks $t_{\text{peak}}$; inset shows the two-block spectrum).
  \textbf{(b,d)} In the stiff mode ($\lambda_{hi}$), the mean $\mu_{hi,t}$ decays rapidly and the variance $v_{hi,t}$ saturates quickly.
  \textbf{(c,e)} In the flat mode ($\lambda_{lo}$), the mean decays slowly while the variance accumulates over long time, producing delayed degradation.
  Notice that the flat-mode terminal variance is substantially larger and dominates late-time reward degradation.}
  \label{fig:toy_modewise}
\end{figure*}

\paragraph{Peak time: a two-timescale characterization.}
The mixture form~\eqref{eq:mixture_exp} implies that peaks arise from competition between modes with different decay rates.
Differentiating (treating $t$ as continuous for analysis) gives
\begin{equation}
  \frac{d}{dt}\Big(\E[\J(\theta_t)]-\J_\infty\Big)
  \;=\;
  -\sum_{\lambda_i>0}\gamma_i A_i e^{-\gamma_i t}.
  \label{eq:dJdt_general}
\end{equation}
An interior maximizer $t_{\mathrm{peak}}>0$ therefore satisfies the implicit condition
\begin{equation}
  \sum_{\lambda_i>0}\gamma_i A_i e^{-\gamma_i t_{\mathrm{peak}}}=0,
  \label{eq:tpeak_implicit}
\end{equation}
which is possible only when the amplitudes $\{A_i\}$ have mixed signs.

\paragraph{Closed form in the two-mode case.}
To make the dependence explicit, consider two distinct eigenvalues
$\lambda_{\mathrm{hi}}>\lambda_{\mathrm{lo}}>0$ with rates
$\gamma_{\mathrm{hi}}=-2\log(1-\alpha\lambda_{\mathrm{hi}})$ and
$\gamma_{\mathrm{lo}}=-2\log(1-\alpha\lambda_{\mathrm{lo}})$, and amplitudes
$A_{\mathrm{hi}}$ and $A_{\mathrm{lo}}$ as in~\eqref{eq:Ai_def}.
Then
\begin{equation}
  \E[\J(\theta_t)]-\J_\infty
  \;=\;
  A_{\mathrm{hi}}e^{-\gamma_{\mathrm{hi}}t}
  +A_{\mathrm{lo}}e^{-\gamma_{\mathrm{lo}}t}.
  \label{eq:two_mode_mix}
\end{equation}
A rise--then--decay peak occurs when $A_{\mathrm{hi}}<0$ and $A_{\mathrm{lo}}>0$
(equivalently, $x_{\mathrm{hi},0}^2>v_{\mathrm{hi},\infty}$ and
$x_{\mathrm{lo},0}^2<v_{\mathrm{lo},\infty}$), and the peak time is
\begin{equation}
  t_{\mathrm{peak}}
  \;=\;
  \frac{1}{\gamma_{\mathrm{hi}}-\gamma_{\mathrm{lo}}}
  \log\!\left(
    \frac{\gamma_{\mathrm{hi}}|A_{\mathrm{hi}}|}
         {\gamma_{\mathrm{lo}}A_{\mathrm{lo}}}
  \right).
  \label{eq:t_peak}
\end{equation}
This expression makes two qualitative dependencies explicit: larger time-scale separation
($\gamma_{\mathrm{hi}}\gg\gamma_{\mathrm{lo}}$) and larger amplitude ratio
($|A_{\mathrm{hi}}|/A_{\mathrm{lo}}$) yield a more pronounced and later peak.

\subsection{Variance--curvature ``spectroscopy'' in the toy model}
\label{app:vcs_toy}

\paragraph{Plateau versus effective noise and curvature-weighted dimension.}
Equation~\eqref{eq:Jinf} implies that, in a locally quadratic and stable regime, the terminal gap scales linearly with $\kappa=\sigma^2/N$:
\begin{equation}
  1-\J_\infty
  = \underbrace{\Bigl(\frac{\alpha}{2}\sum_{\lambda_i>0}\frac{1}{2-\alpha\lambda_i}\Bigr)}_{\text{curvature functional}}
  \cdot \kappa.
  \label{eq:slope_linear}
\end{equation}
This motivates the curvature-weighted effective dimension
\begin{equation}
  d_{\mathrm{eff}}(\alpha)\;\defeq\;2\sum_{\lambda_i>0}\frac{1}{2-\alpha\lambda_i},
  \qquad
  \Rightarrow\qquad
  1-\J_\infty = \frac{\alpha}{4}\,d_{\mathrm{eff}}(\alpha)\cdot \kappa.
  \label{eq:deff}
\end{equation}
For $\alpha\lambda_i\ll 1$, $d_{\mathrm{eff}}(\alpha)\approx \mathrm{rank}(C)$; at finite step size it smoothly reweights directions by curvature through $(2-\alpha\lambda_i)^{-1}$.
Thus, in the toy quadratic setting, a slope fit of $(1-\J_\infty)$ versus $\kappa$ recovers a scalar summary of local curvature complexity.

\begin{figure*}[t]
  \centering
  \includegraphics[width=0.7\textwidth]{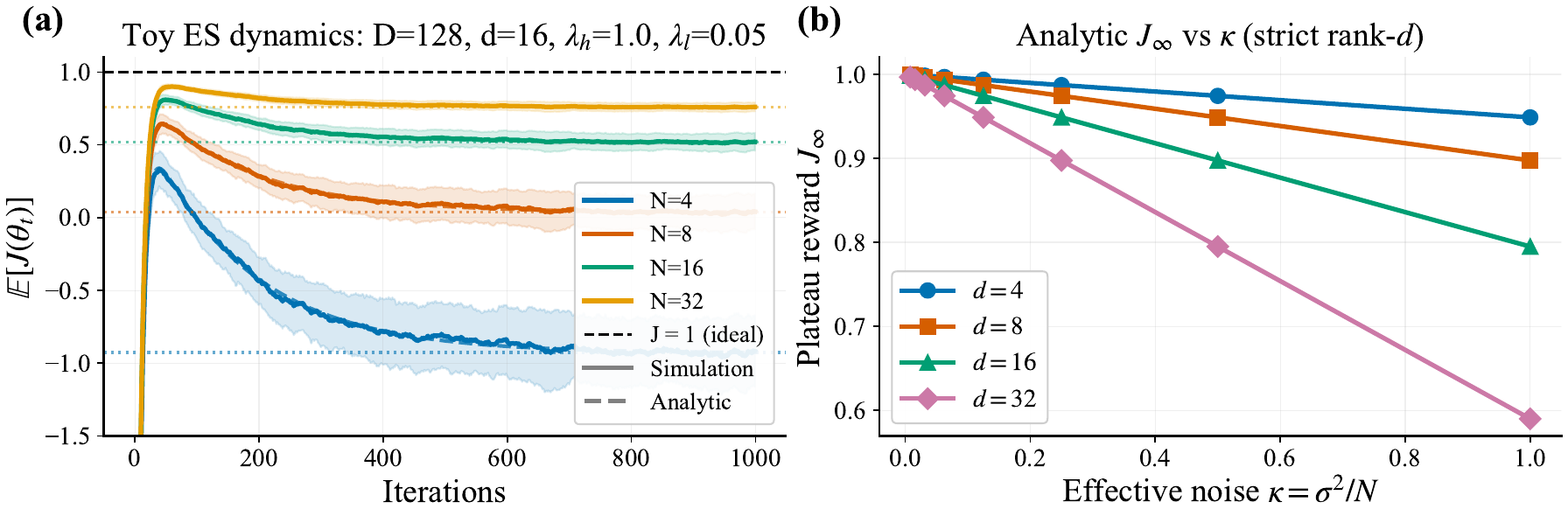}
  \caption{\textbf{Toy dynamics and plateau ``spectroscopy'' via effective noise $\kappa=\sigma^2/N$.}
  \textbf{(a)} Expected reward trajectories for different populations $N$ (noise levels) on the two-block quadratic landscape, comparing Monte Carlo ES simulations to the analytic OU prediction; larger $N$ reduces effective noise and raises the terminal plateau.
  \textbf{(b)} Analytic terminal plateau $J_\infty$ versus effective noise $\kappa$ for strict rank-$d$ curvature.
  The approximately linear dependence predicted by Eq.~\eqref{eq:slope_linear} has a slope that grows with $d$, showing that plateau-vs-noise measurements recover a curvature-weighted effective dimension in the toy setting.}
  \label{fig:toy_spectroscopy}
\end{figure*}

\paragraph{Slope fitting as ``spectroscopy'' in the toy setting.}
Figure~\ref{fig:toy_spectroscopy} illustrates this viewpoint.
Varying population size $N$ changes $\kappa$ and produces different terminal plateaus in the reward trajectory (panel a).
Plotting the analytic plateau $J_\infty$ against $\kappa$ yields an approximately linear dependence whose slope increases with the number of curvature-active directions in a strict rank-$d$ spectrum (panel b).
In other words, higher effective dimension implies greater sensitivity of the noise floor to stochasticity.
This motivates the checkpointed noise-floor probing method introduced next (Section~\ref{app:clss_slq}).

\section{From Toy Slope Laws to Practical Curvature Probes: CLSS vs.\ SLQ}
\label{app:clss_slq}

The quadratic OU analysis in Appendix~\ref{app:toy_derivation} yields a concrete measurement principle: in a locally stable region, the terminal ``noise-floor'' gap scales approximately linearly with the effective noise level $\kappa=\sigma^2/N$ (Eq.~\eqref{eq:slope_linear}), and the slope can be mapped to a curvature-weighted effective dimension (Eq.~\eqref{eq:deff}).
In the toy setting this relation is exact and enables a simple slope fit to recover $d_{\mathrm{eff}}$ (Fig.~\ref{fig:toy_spectroscopy}).
This motivates a practical idea for black-box or reward-defined objectives where Hessian-vector products may be unavailable: estimate local geometry by \emph{checkpointed} noise-floor probing and slope fitting.
We refer to this as \emph{Checkpointed Local Slope Spectroscopy (CLSS)}.
We then contrast CLSS with Stochastic Lanczos Quadrature (SLQ), a standard matvec-based spectrum probe.

\subsection{Checkpointed Local Slope Spectroscopy (CLSS)}
Given a checkpoint $\theta^\star$, CLSS runs short, local probe trajectories at several controllable noise levels (e.g., varying $N$ at fixed $\sigma$ in ES), estimates a plateau proxy for each noise level, and fits the small-noise slope of the plateau gap versus $\kappa=\sigma^2/N$.
In the quadratic regime, this slope maps to an effective dimension via Eq.~\eqref{eq:slope_linear}--\eqref{eq:deff}. The algorithmic description of CLSS is given in Algorithm~\ref{alg:clss_concise}. 
\begin{algorithm}[t]
\caption{Checkpointed Local Slope Spectroscopy (CLSS)}
\label{alg:clss_concise}
\KwIn{
checkpoint $\theta^\star$; perturbation scale $\sigma$; candidate step sizes $\mathcal A$;
population sizes $\mathcal N$; probe horizon $T$; tail window $w$; \#seeds $R$;\\
locality metric $\mathrm{Loc}(\theta,\theta^\star)$ and threshold $\tau_{\mathrm{loc}}$;
tail-settling tolerance $\tau_{\mathrm{stat}}$; minimum valid seeds $R_{\min}$.
}
\KwOut{$\widehat d_{\mathrm{eff}}(\alpha;\theta^\star)$ with CI, or \texttt{FAIL}.}

\BlankLine
\ForEach{$\alpha \in \mathcal A$}{
  \tcp{(1) Local probes at fixed $\alpha,\sigma$ and varying population $N$ (noise $\kappa=\sigma^2/N$)}
  \ForEach{$N \in \mathcal N$}{
    $\mathcal V \gets \emptyset$ \tcp*{valid seeds}
    \For{$r=1$ \KwTo $R$}{
      Run $T$ probe steps from $\theta^\star$ with $(\alpha,\sigma,N)$ producing rewards $\{J_t\}_{t=1}^T$ and states $\{\theta_t\}_{t=1}^T$\;
      \If{$\max_{t\le T}\mathrm{Loc}(\theta_t,\theta^\star)>\tau_{\mathrm{loc}}$}{\textbf{continue}}
      $\bar J_1 \gets \frac1w\sum_{t=T-w+1}^{T} J_t$\;
      $\bar J_0 \gets \frac1w\sum_{t=T-2w+1}^{T-w} J_t$\;
      \If{$|\bar J_1-\bar J_0|>\tau_{\mathrm{stat}}$}{\textbf{continue}}
      $\mathcal V \gets \mathcal V \cup \{\bar J_1\}$\;
    }
    \If{$|\mathcal V|<R_{\min}$}{mark $N$ invalid; \textbf{continue}}
    $\widehat J_\infty(N) \gets \mathrm{mean}(\mathcal V)$\;
  }

  \tcp{(2) Fit small-noise slope of plateau gap vs $\kappa=\sigma^2/N$}
  Let $N_{\max}$ be the largest valid $N$ and set $\widehat J_{\mathrm{ref}}\gets \widehat J_\infty(N_{\max})$\;
  For each valid $N$, set $g(N)\gets \widehat J_{\mathrm{ref}}-\widehat J_\infty(N)$\;
  Fit $g(N)\approx S_\alpha\cdot(\sigma^2/N)+b$ using the largest few valid $N$ (small-noise regime)\;
  Output $\widehat d_{\mathrm{eff}}(\alpha;\theta^\star)\gets \frac{4}{\alpha}\,S_\alpha$ \tcp*{by Eq.~\eqref{eq:deff}}
  \If{fit residuals small and acceptance rates high}{\KwRet $\widehat d_{\mathrm{eff}}(\alpha;\theta^\star)$}
}
\KwRet \texttt{FAIL}.
\end{algorithm}
\paragraph{Practical limitations.}
CLSS is conceptually clean in the quadratic regime but can be difficult to apply reliably in real LLM fine-tuning.
It requires (i) a \emph{local} probe regime (trajectories must remain near $\theta^\star$), and (ii) a \emph{measurable} plateau proxy over the probe horizon.
In practice, nonconvex drift across regions, long transients, and bounded/truncated rewards (e.g., accuracy in $[0,1]$) can obscure plateaus and inflate uncertainty.
Moreover, effective noise is often anisotropic and state-dependent (particularly for policy-gradient methods), weakening a direct mapping from a scalar slope to a single curvature functional.
For these reasons, in the main text we use the slope law primarily as a \emph{mechanistic explanation} and rely on more robust operational probes (e.g., best-of-$N$ accessibility) that do not require clean plateau observation.

\subsection{Stochastic Lanczos Quadrature (SLQ): matvec-based curvature probing}
SLQ estimates trace functionals and (optionally) spectral densities of a symmetric operator $A$ using only matrix--vector products $v\mapsto Av$.
It combines Hutchinson trace estimation with Lanczos tridiagonalization and Gaussian quadrature to approximate quantities of the form $\mathrm{tr}(f(A))$ with $\mathcal{O}(s m)$ matvecs, where $s$ is the number of random probes and $m$ the number of Lanczos steps~\citep{hutchinson1990trace,ubaru2017slq}.
Tooling such as \textsc{PyHessian} implements related workflows for estimating top eigenvalues, trace, and smoothed spectral densities from Hessian--vector products~\citep{yao2019pyhessian,ghorbani2019hessian_density}.
When a reliable differentiable surrogate loss is available and matvecs are stable, SLQ provides rich spectral diagnostics (bulk/outliers, sharp subspaces) that complement our dynamics-based analysis. The algorithmic description of SLQ is given in Algorithm~\ref{alg:slq}. 

\begin{algorithm}[t]
\caption{Stochastic Lanczos Quadrature (SLQ) for $\mathrm{tr}(f(A))$}
\label{alg:slq}
\KwIn{
symmetric matrix/operator $A$ (accessed only via matvec $v\mapsto Av$);
scalar function $f$; \#probes $s$; Lanczos steps $m$.
}
\KwOut{
estimate $\widehat{\mathrm{tr}(f(A))}$ and (optionally) a standard error.
}

\BlankLine
\tcp{Hutchinson trace estimator: $\mathrm{tr}(f(A)) = \mathbb{E}[z^\top f(A) z]$ for $\mathbb{E}[zz^\top]=I$}
\For{$j=1$ \KwTo $s$}{
  Sample probe $z_j \in \{\pm1\}^n$ i.i.d.\ Rademacher (or $z_j\sim\mathcal N(0,I)$).\\
  Normalize $q_1 \gets z_j / \|z_j\|_2$.\\

  \tcp{Lanczos tridiagonalization using only matvecs with $A$}
  Run $m$-step Lanczos on $A$ with start $q_1$ to obtain tridiagonal $T_j\in\mathbb{R}^{m\times m}$.\\

  \tcp{Gaussian quadrature: $z^\top f(A) z \approx \|z\|^2 e_1^\top f(T)e_1$}
  Compute eigendecomposition $T_j = V_j \mathrm{diag}(\theta_j) V_j^\top$.\\
  Set weights $w_{jk} \gets (V_j)_{1k}^2$ for $k=1,\dots,m$.\\
  Estimate quadratic form:
  \begin{equation}
    \widehat{\tau}_j \;\gets\; \|z_j\|_2^2 \sum_{k=1}^m w_{jk}\, f(\theta_{jk}).
  \end{equation}
}
\BlankLine
\begin{equation}
  \widehat{\mathrm{tr}(f(A))} \;\gets\; \frac{1}{s}\sum_{j=1}^s \widehat{\tau}_j,
  \qquad
  \widehat{\mathrm{SE}} \;\gets\; \sqrt{\frac{\mathrm{Var}(\{\widehat{\tau}_j\}_{j=1}^s)}{s}} \;\;(\text{optional}).
\end{equation}
\KwRet $\widehat{\mathrm{tr}(f(A))}$ (and $\widehat{\mathrm{SE}}$).
\end{algorithm}

\subsection{When to use what}
\begin{itemize}[leftmargin=*]
\item \textbf{Use SLQ/\textsc{PyHessian} when matvecs are available.}
If you can compute stable matvecs for a symmetric curvature operator tied to a differentiable surrogate (Hessian/GGN/Fisher), SLQ yields detailed spectral structure and top-eigenspace information.
\item \textbf{Use CLSS when only objective evaluations are available and plateaus are measurable.}
CLSS is applicable to black-box rewards and algorithm-native noise control (e.g., varying $N$ in ES), but only to the extent that local probes remain near a checkpoint and produce a reliably measurable plateau.
\item \textbf{In our LLM setting, favor operational probes.}
Because plateau estimation is often unreliable under bounded/truncated rewards and nonstationarity, we do not use CLSS as a primary measurement and instead emphasize robust probes (best-of-$N$) and dynamical signatures (rise--then--decay) in the main paper.
\end{itemize}

\begin{table}[t]
\centering
\small
\setlength{\tabcolsep}{4pt}
\renewcommand{\arraystretch}{1.12}
\begin{tabularx}{\linewidth}{
  l
  >{\raggedright\arraybackslash}X
  >{\raggedright\arraybackslash}X
}
\toprule
\textbf{Method} & \textbf{What it needs} & \textbf{What it gives} \\
\midrule
\textbf{SLQ / \textsc{PyHessian}} &
Matvec access $v\mapsto Av$ for a symmetric curvature operator (typically via differentiable surrogate) &
Rich spectral diagnostics: top eigenvalues/eigenspace, trace functionals, smoothed spectral density \\
\midrule
\textbf{CLSS (slope probe)} &
Only objective evaluations under the actual optimizer; controllable noise $\kappa=\sigma^2/N$; a measurable local plateau &
Scalar slope-based summary interpretable as curvature-weighted effective dimension in the quadratic regime \\
\bottomrule
\end{tabularx}
\caption{\textbf{High-level comparison.} SLQ provides detailed spectral information when matvecs are available; CLSS is inspired by the toy slope law and can be used with black-box rewards, but requires reliable local plateau estimation.}
\label{tab:slq_clss_summary}
\end{table}

\section{Anisotropic Noise and the Lyapunov Equation}
\label{app:anisotropic_lyapunov}

The quadratic OU toy model in Appendix~\ref{app:toy_derivation} assumed isotropic noise for clarity, which diagonalizes neatly in the eigenbasis of the curvature matrix.
In practice, stochastic learning rules inject \emph{anisotropic} and often state-dependent noise.
This appendix summarizes the corresponding linearized theory, which shows that the same geometry--variance mechanism persists even when modes do not decouple: curvature still sets relaxation, while the noise covariance determines how variance accumulates across directions.

\subsection{Linearized dynamics with anisotropic noise}
Near a critical point $\theta^\star$, a general linearized update takes the form
\begin{equation}
  x_{t+1} = (I-\alpha H)x_t + \alpha \eta_t,\qquad
  \mathbb{E}[\eta_t]=0,\;\;\mathrm{Cov}(\eta_t)=\Sigma,
  \label{eq:lin_update_appendix}
\end{equation}
where $x_t=\theta_t-\theta^\star$ and $H$ is the local curvature matrix (for reward maximization, $H=-C\succeq 0$ in our sign convention).
Let $V_t=\mathbb{E}[x_t x_t^\top]$.
Then
\begin{equation}
  V_{t+1} = (I-\alpha H)V_t(I-\alpha H)^\top + \alpha^2 \Sigma.
  \label{eq:discrete_lyapunov}
\end{equation}
At stationarity ($V_{t+1}=V_t=V$), this is the \emph{discrete Lyapunov equation}.
For small $\alpha$, expanding yields the continuous-time OU covariance balance
\begin{equation}
  HV + VH^\top \approx \alpha \Sigma,
  \label{eq:cont_lyapunov}
\end{equation}
matching standard OU/SDE approximations of constant-step stochastic optimization~\citep{mandt2017sgd_as_bayes,li2017stochastic_modified_equations}.

\paragraph{How the plateau depends on anisotropic noise.}
In the quadratic approximation $\J(\theta^\star+x)\approx \J(\theta^\star)-\tfrac{1}{2}x^\top Hx$, the expected performance gap depends on the stationary second moment:
\begin{equation}
  \J(\theta^\star)-\mathbb{E}[\J(\theta^\star+x)]
  \;\approx\;
  \tfrac{1}{2}\,\mathrm{tr}(H V).
  \label{eq:gap_traceHV}
\end{equation}
Thus, even when $H$ is low-dimensional in curvature (few stiff modes), the \emph{effective noise floor} depends on how $\Sigma$ injects variance into those modes.
Compared to the isotropic case (Appendix~\ref{app:toy_derivation}), anisotropy can (i) selectively amplify variance in particular curvature directions, (ii) rotate variance into stiff directions through off-diagonal structure, and (iii) change the apparent ``effective dimension'' by reweighting modes through $\Sigma$ rather than through $H$ alone.

\paragraph{Stability and the OU regime.}
A stationary $V$ exists only when the linear drift is stable: the spectral radius of $(I-\alpha H)$ must be $<1$ (in the reward-maximization convention, this corresponds to $H\succeq 0$ on the relevant subspace and $\alpha$ sufficiently small).
When this holds, \eqref{eq:discrete_lyapunov} provides a precise statement of the ``variance accumulation'' mechanism in the presence of anisotropy: the curvature $H$ sets contraction, while $\Sigma$ sets variance injection.

\subsection{Policy-gradient / GRPO noise structure}
\label{app:pg_noise}

Policy-gradient methods provide a concrete example of anisotropic noise.
Let $\mathcal{L}(\theta)$ denote a differentiable surrogate loss used for optimization (e.g., a GRPO/PPO-style objective).
A generic stochastic gradient update has the form
\begin{equation}
\theta_{t+1}=\theta_t-\alpha\,\widehat g_t,\qquad \widehat g_t=\nabla \mathcal{L}(\theta_t)+\zeta_t,
\end{equation}
where $\zeta_t$ is stochastic gradient noise induced by sampling trajectories/tokens and minibatches.
Under local linearization around $\theta^\star$, this yields \eqref{eq:lin_update_appendix} with $H=\nabla^2\mathcal{L}(\theta^\star)$ and $\Sigma\approx \mathrm{Cov}(\zeta_t)$.

\paragraph{Score-function form and Fisher-like covariance.}
For a standard policy gradient estimator (for notational simplicity suppressing state conditioning),
\begin{equation}
\widehat g = -\frac{1}{B}\sum_{b=1}^B \widehat A_b \,\nabla_\theta \log \pi_\theta(a_b),
\end{equation}
where $\widehat A_b$ is an advantage-like scalar and $B$ is the batch size.
Conditioned on $\theta$, the covariance of $\widehat g$ can be decomposed into (i) a Fisher-like term from $\nabla\log\pi$ and (ii) scaling/mixing induced by the random advantages:
\begin{equation}
\Sigma \;\approx\; \mathrm{Cov}(\widehat g)\;=\;\frac{1}{B}\Big(\mathbb{E}[\widehat A^2\,(\nabla\log\pi)(\nabla\log\pi)^\top] - \mathbb{E}[\widehat A\,\nabla\log\pi]\,\mathbb{E}[\widehat A\,\nabla\log\pi]^\top\Big).
\end{equation}
Even when $H$ is low-dimensional in curvature, the effective noise floor is determined by how this anisotropic $\Sigma$ projects onto curvature-active directions through the Lyapunov equation~\eqref{eq:cont_lyapunov}.

\paragraph{Effect of GRPO group normalization.}
GRPO introduces a \emph{group-relative} normalization that centers (and often scales) advantages within groups.
At a high level, centering acts like a control variate that removes components of noise aligned with the within-group mean, while scaling by a within-group standard deviation can change the effective noise magnitude and induce correlations across samples within a group.
Consequently, GRPO changes $\Sigma$ in a structured, data-dependent manner rather than simply reducing variance uniformly.
This provides a natural explanation for why non-monotonic reward dynamics can appear in GRPO as well as ES (Fig.~\ref{fig:exp_nonmono}): even if the mean update direction improves performance early, anisotropic noise interacting with heterogeneous curvature can drive a variance-dominated late regime.

\paragraph{Empirical proxies for $\Sigma$.}
In practice, $\Sigma$ can be estimated (approximately) by minibatch gradient covariance:
\begin{equation}
\widehat\Sigma \;=\; \frac{1}{B-1}\sum_{b=1}^B (\widehat g_b-\bar g)(\widehat g_b-\bar g)^\top,
\end{equation}
or by low-rank projections (e.g., Hutchinson probing or subspace restrictions) when full matrices are infeasible.
Combined with top-eigenspace estimates of $H$ (when available), the Lyapunov relation provides a principled way to predict which directions will accumulate variance and when a variance-dominated regime is expected.

\section{Beyond Quadratic: Saddles and Negative Curvature}
\label{app:beyond_quadratic}

The OU analysis in the main text and Appendix~\ref{app:toy_derivation} describes \emph{within-basin} dynamics around a locally stable region (a local maximizer for reward, or minimizer for loss).
Real fine-tuning landscapes are nonconvex and nonstationary, and two additional effects can complicate plateau-based reasoning and amplify non-monotonicity.

\paragraph{Saddles and negative curvature.}
OU stationarity requires stable drift.
Near strict saddles, negative-curvature directions make the linear drift unstable (e.g., eigenvalues of $(I-\alpha H)$ have magnitude $>1$), so a stationary local covariance does not exist.
In practice, stochasticity can help escape saddles; perturbed gradient methods escape strict saddles efficiently under suitable conditions~\citep{jin2017escape_saddles}.
In such regions, apparent ``plateaus'' can be transient and slope-based noise-floor estimation can be misleading.

\paragraph{Basin-to-basin transitions and nonstationarity.}
Even near locally stable regions, finite-step stochastic updates can drift across regions of the landscape, especially along flat directions.
This can invalidate single-basin plateau assumptions and produce additional non-monotonicity beyond the OU mechanism.
For RL-style fine-tuning, further nonstationarity arises because the effective objective can change with the policy distribution, sampling temperature, KL penalties, or group normalization.
Qualitatively, these effects are consistent with the observation that GRPO trajectories can also exhibit peak--then--decay behavior (Fig.~\ref{fig:exp_nonmono}), even though GRPO optimizes a differentiable surrogate: both the surrogate geometry and its noise structure can evolve during training.

\paragraph{Implications for our measurements.}
These considerations motivate two design choices in the main paper.
First, we treat rise--then--decay as a \emph{diagnostic} of heterogeneous curvature and variance coupling, rather than as evidence of a literal stationary OU regime throughout training.
Second, we emphasize operational probes (e.g., best-of-$N$ accessibility) that do not require a clean, long-horizon plateau in a single basin, and we relegate plateau-based slope-fitting ideas to controlled toy settings and appendix discussion.

\section{Metastability in a Double-Well: Escape Times and Hopping Criteria}
\label{app:double_well}

The quadratic OU analysis describes within-basin behavior near a locally stable region.
To understand when stochastic optimization transitions between basins (``hopping''), we
summarize the canonical metastable case: overdamped Langevin dynamics in a double-well.
The key takeaway is that escape times depend \emph{exponentially} on a barrier-to-noise ratio,
so trajectories can appear either (i) effectively trapped or (ii) rapidly delocalized,
with a narrow intermediate regime of rare hopping.

\subsection{From noisy gradient steps to an overdamped Langevin diffusion}
Consider one-dimensional noisy gradient descent on a loss $L(x)$,
\begin{equation}
  x_{k+1} \;=\; x_k - \alpha L'(x_k) + \alpha\,\xi_k,
  \qquad \xi_k \sim \mathcal{N}(0,\sigma^2/N).
  \label{eq:sgd_1d}
\end{equation}
For small step size $\alpha$,~\eqref{eq:sgd_1d} is the Euler--Maruyama discretization of the
overdamped Langevin SDE
\begin{equation}
  d x_t \;=\; -L'(x_t)\,dt + \sqrt{2\varepsilon}\,dW_t,
  \label{eq:langevin}
\end{equation}
with effective noise intensity $\varepsilon$ obtained by matching per-step variances.
Identifying $dt=\alpha$ and matching $2\varepsilon\,dt \approx \alpha^2(\sigma^2/N)$ yields
\begin{equation}
  \varepsilon \;\approx\; \frac{\alpha\sigma^2}{2N}
  \;=\; \frac{\alpha}{2}\,\kappa,
  \label{eq:epsilon}
\end{equation}
where $\kappa=\sigma^2/N$ is the effective ES noise scale used throughout the paper.

\subsection{Eyring--Kramers escape time in 1D}
Let $x_-$ be a local minimum of $L$ and let $z$ be the adjacent saddle (in 1D, the local maximum
separating the well). Define the barrier height $\Delta L \defeq L(z)-L(x_-)$.
In the small-noise regime $\varepsilon\to 0$, the mean first exit time obeys the
Eyring--Kramers law
\begin{equation}
  \E[\tau_{\mathrm{esc}}]
  \;\approx\;
  \frac{2\pi}{\sqrt{L''(x_-)\,|L''(z)|}}\,
  \exp\!\Big(\frac{\Delta L}{\varepsilon}\Big),
  \label{eq:ek_1d}
\end{equation}
up to lower-order corrections~\citep{hanggi1990kramers,lelievre2025eyring}.
Since one discrete step corresponds to $dt=\alpha$, the expected number of iterations to escape is
\begin{equation}
  \E[K_{\mathrm{esc}}]
  \;\approx\;
  \frac{1}{\alpha}\,\E[\tau_{\mathrm{esc}}]
  \;\approx\;
  \frac{2\pi}{\alpha\sqrt{L''(x_-)\,|L''(z)|}}\,
  \exp\!\Big(\frac{\Delta L}{\varepsilon}\Big).
  \label{eq:ek_iters}
\end{equation}
Consequently, the probability of at least one hop within $T$ iterations is approximately
\begin{equation}
  \Pr(\text{hop by }T)
  \;\approx\;
  1-\exp\!\Big(-\frac{T}{\E[K_{\mathrm{esc}}]}\Big)
  \;\approx\;
  \frac{T}{\E[K_{\mathrm{esc}}]}
  \quad \text{when } T\ll \E[K_{\mathrm{esc}}].
  \label{eq:hop_prob}
\end{equation}

\subsection{Specialization to the quartic double-well}
For the standard quartic double-well
\begin{equation}
  L(x) \;=\; \frac{\lambda_{\mathrm{dw}}}{4}\,(x^2-a^2)^2,
  \label{eq:dw_loss}
\end{equation}
the minima are at $x_\pm=\pm a$ and the saddle is at $z=0$.
The barrier height is
\begin{equation}
  \Delta L \;=\; L(0)-L(-a) \;=\; \frac{\lambda_{\mathrm{dw}}}{4}\,a^4,
  \label{eq:dw_barrier}
\end{equation}
and the curvatures are
\begin{equation}
  L''(\pm a)=2\lambda_{\mathrm{dw}}a^2,
  \qquad
  L''(0)=-\lambda_{\mathrm{dw}}a^2.
  \label{eq:dw_curvatures}
\end{equation}
Plugging~\eqref{eq:dw_barrier}--\eqref{eq:dw_curvatures} into~\eqref{eq:ek_iters} yields
\begin{equation}
  \E[K_{\mathrm{esc}}]
  \;\approx\;
  \frac{2\pi}{\alpha\,\lambda_{\mathrm{dw}}a^2\sqrt{2}}\,
  \exp\!\Big(\frac{\Delta L}{\varepsilon}\Big)
  \;=\;
  \frac{2\pi}{\alpha\,\lambda_{\mathrm{dw}}a^2\sqrt{2}}\,
  \exp\!\Big(\frac{\lambda_{\mathrm{dw}}a^4}{2}\cdot\frac{N}{\alpha\sigma^2}\Big).
  \label{eq:dw_escape_iters}
\end{equation}

\subsection{A practical hopping criterion and regimes}
\begin{figure*}[t]
  \centering
  \includegraphics[width=0.9\textwidth]{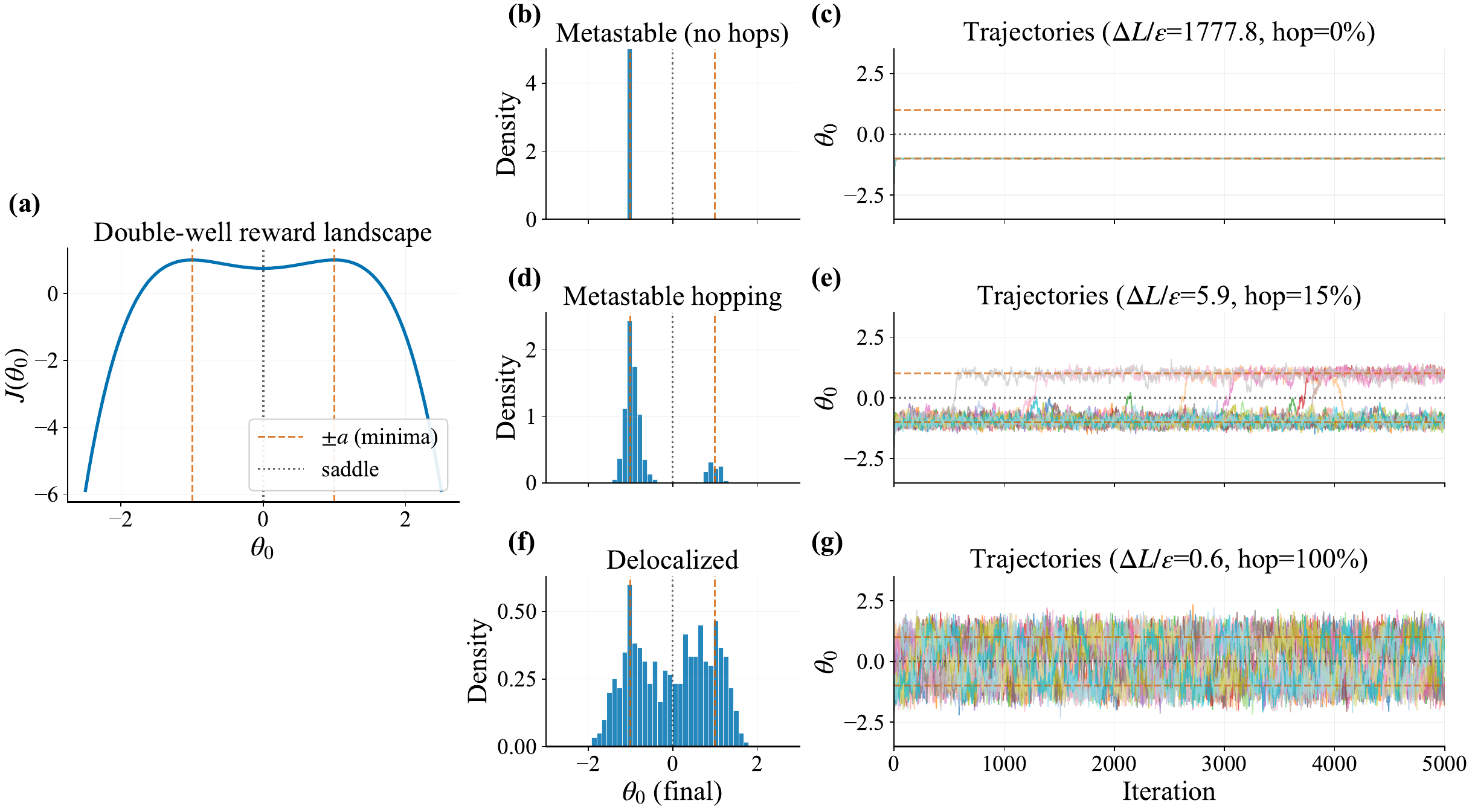}
  \caption{\textbf{Metastability and basin hopping in a double-well under stochastic updates.}
  \textbf{(a)} One-dimensional double-well reward landscape $J(\theta_0)$ (with other coordinates set to zero), with minima at $\pm a$ and a saddle near $0$.
  \textbf{(b,d,f)} Histograms of $\theta_0$ at the final iteration across runs.
  \textbf{(c,e,g)} Example trajectories of $\theta_0(t)$ across runs.
  The three rows illustrate the regimes predicted by the barrier-to-noise ratio $\Delta L/\varepsilon$ (Eq.~\eqref{eq:barrier_over_temp}): 
  \textbf{top:} metastable confinement (no hops), 
  \textbf{middle:} rare basin hopping, and 
  \textbf{bottom:} delocalized behavior with frequent transitions.
  ``hop\%'' denotes the fraction of runs that switch wells at least once under a simple hysteresis-based definition (Appendix~\ref{app:double_well}).}
  \label{fig:double_well_regimes}
\end{figure*}

Escape is controlled by the dimensionless barrier-to-noise ratio
\begin{equation}
  \frac{\Delta L}{\varepsilon}
  \;=\;
  \frac{\lambda_{\mathrm{dw}}a^4/4}{\alpha\sigma^2/(2N)}
  \;=\;
  \frac{\lambda_{\mathrm{dw}}a^4}{2}\cdot\frac{N}{\alpha\sigma^2}.
  \label{eq:barrier_over_temp}
\end{equation}
Because $\E[K_{\mathrm{esc}}]$ depends \emph{exponentially} on~\eqref{eq:barrier_over_temp},
a coarse threshold for observing at least one hop within $T$ iterations is
\begin{equation}
  \frac{\Delta L}{\varepsilon} \;\approx\; \log T,
  \label{eq:hopping_threshold}
\end{equation}
up to logarithmic corrections from the prefactor.
This yields three qualitative regimes:
\begin{itemize}[leftmargin=*]
  \item \textbf{Metastable (no hops):} $\Delta L/\varepsilon \gg \log T$.
  \item \textbf{Metastable hopping:} $\Delta L/\varepsilon \approx O(\log T)$.
  \item \textbf{Delocalized:} $\Delta L/\varepsilon \lesssim 1$.
\end{itemize}
Figure~\ref{fig:double_well_regimes} visualizes these three regimes in simulation.
As $\Delta L/\varepsilon$ decreases, trajectories transition from remaining confined to one well (top), to occasional barrier crossings over the horizon (middle), and finally to frequent crossings that wash out basin localization (bottom).
This sharp qualitative transition over finite horizons reflects the exponential sensitivity of escape times to the barrier-to-noise ratio in Eq.~\eqref{eq:ek_iters}.


\subsection{Local ``water level'' within a well}
Within one well (e.g., near $x=-a$), $L$ is approximately quadratic with curvature
$\kappa_{\mathrm{well}}=L''(-a)=2\lambda_{\mathrm{dw}}a^2$, yielding an OU approximation
with stationary variance
\begin{equation}
  \mathrm{Var}(x)\;\approx\;\frac{\varepsilon}{\kappa_{\mathrm{well}}}
  \;=\;
  \frac{\alpha\sigma^2}{4N\,\lambda_{\mathrm{dw}}a^2},
  \label{eq:local_var}
\end{equation}
making explicit how local curvature suppresses fluctuations within a basin.

\paragraph{Relevance to the main analysis (from within-basin OU to basin hopping).}
The quadratic OU analysis in Appendix~\ref{app:toy_derivation} isolates a \emph{within-basin} mechanism: near a locally stable region, fixed stochasticity induces a noise-controlled plateau and, with heterogeneous curvature, can produce rise--then--decay without requiring nonconvex pathologies.
The double-well analysis in Appendix~\ref{app:double_well} complements this picture by characterizing when stochastic learning leaves a basin altogether.
Its key message is that basin transitions are controlled by an exponentially sensitive barrier-to-noise ratio (Eq.~\eqref{eq:barrier_over_temp}), yielding three regimes over a finite training horizon $T$: effectively no hops, rare hops, or delocalization.

In a high-dimensional nonconvex fine-tuning landscape, local neighborhoods can be viewed as collections of basins separated by saddles of varying heights, and different stochastic learning rules induce different effective noise intensities and directions.
Consequently, basin-hopping need not be ubiquitous: even modest changes in the effective noise level (through $\alpha$, $\sigma^2/N$, temperature, batch size, or group size) can move training across the hopping threshold~\eqref{eq:hopping_threshold} for some barriers but not others.
Operationally, this provides a second (non-exclusive) source of late-stage non-monotonicity beyond within-basin variance accumulation: if stochasticity is large enough to cross low barriers along weakly constrained directions, training may drift between nearby basins, leading to additional variability or degradation in reward.
Conversely, when $\Delta L/\varepsilon \gg \log T$ for the relevant barriers, training remains effectively confined and the within-basin OU mechanism is the appropriate local description.
This perspective motivates treating non-monotonic dynamics as potentially arising from both (i) within-basin variance--curvature effects and (ii) finite-horizon basin-hopping, with their relative importance governed by the same effective noise scale.

\section{Additional Rise--Then--Decay Results: Countdown}
\label{sec:appendix_countdown}

\begin{figure}[t]
  \centering
  \includegraphics[width=0.85\linewidth]{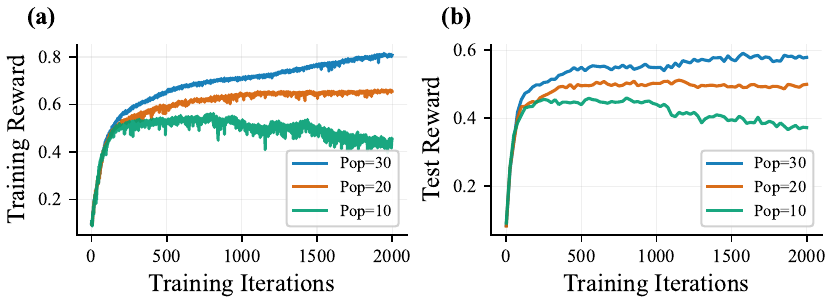}
  \caption{\textbf{Rise--then--decay behavior on the Countdown task.}
  Training (left) and held-out test (right) reward trajectories for ES fine-tuning on the Countdown arithmetic task under fixed hyperparameters, shown for population sizes $N\in\{10,20,30\}$.
  Across all population sizes, reward improves rapidly from the pretrained checkpoint, reaches a peak, and subsequently declines toward a lower value.
  Larger populations delay the onset and reduce the magnitude of decay but do not eliminate the non-monotonic behavior.
  These dynamics closely mirror those observed on GSM8K, ARC-C, and WinoGrande, indicating that rise--then--decay is a robust consequence of stochastic fine-tuning interacting with anisotropic curvature rather than a task-specific artifact.}
  \label{fig:countdown_nonmono}
\end{figure}

We provide additional evidence for non-monotonic fine-tuning dynamics on the Countdown task, a structured arithmetic reasoning benchmark distinct from GSM8K, ARC-C, and WinoGrande.
Figure~\ref{fig:countdown_nonmono} shows ES training and test reward trajectories under fixed hyperparameters for population sizes $N\in\{10,20,30\}$.

Consistent with the main experiments, all runs exhibit a pronounced rise--then--decay pattern.
Reward initially increases as stochastic updates exploit high-curvature (stiff) directions associated with rapid improvement, but later decreases as improvement along these directions saturates and stochastic drift accumulates along weakly curved (flat) directions.
Increasing the population size delays the onset of decay and raises peak reward, reflecting reduced sampling variance, but does not qualitatively alter the trajectory shape.

These results reinforce two key conclusions.
First, non-monotonic reward trajectories are not tied to a specific task, evaluation protocol, or benchmark family, but arise generically under fixed stochasticity.
Second, the qualitative dependence on population size matches the predictions of the geometry--variance framework: stochasticity interacts with anisotropic curvature to produce early gains followed by late-time degradation once signal along stiff directions is exhausted.
Together with the main results, the Countdown experiments support the view that rise--then--decay is a structural consequence of fine-tuning landscapes rather than an artifact of a particular dataset or algorithmic detail.

\section{Experimental Details: ES Reward-Probe}
\label{app:exp_details_es_probe}

This appendix describes the ES reward-probe protocol used to generate the scaling figures in the main text (e.g., Fig.~\ref{fig:es_scaling_combined}) and the related appendix analyses. The goal of these probes is to characterize reward changes under random weight perturbations at controlled noise scale and to estimate extreme-value quantities (best-of-$N$) in a manner that cleanly separates \emph{perturbation randomness} (algorithmic) from \emph{evaluation-set randomness} (measurement).

\subsection{Models and inference}
We evaluate instruction-tuned models from the Qwen2.5-Instruct family:
\begin{center}
\begin{tabular}{ll}
\toprule
\textbf{Model size} & \textbf{HuggingFace model ID} \\
\midrule
0.5B & \texttt{Qwen/Qwen2.5-0.5B-Instruct} \\
1.5B & \texttt{Qwen/Qwen2.5-1.5B-Instruct} \\
3B & \texttt{Qwen/Qwen2.5-3B-Instruct} \\
7B & \texttt{Qwen/Qwen2.5-7B-Instruct} \\
\bottomrule
\end{tabular}
\end{center}
Models are loaded in \texttt{bfloat16}. Inference uses HuggingFace Transformers with vLLM as the generation backend; for consistency we use eager attention mode (flash attention disabled). Decoding is greedy (temperature 0.0) with a maximum of 512 new tokens. A fixed generation seed (42) is used across candidates to ensure deterministic prompt-to-output mapping given parameters.

\subsection{Tasks, datasets, and rewards}
We probe three tasks using the training splits of standard datasets and binary per-prompt rewards:
\begin{center}
\begin{tabular}{llll}
\toprule
\textbf{Task} & \textbf{Dataset} & \textbf{Split} & \textbf{Reward} \\
\midrule
GSM8K & \texttt{gsm8k/main} & train & 1 iff final number matches GT \\
ARC-C & \texttt{allenai/ai2\_arc} (Challenge) & train & 1 iff correct A/B/C/D \\
WinoGrande & \texttt{allenai/winogrande} (\texttt{winogrande\_xl}) & train & 1 iff correct A/B \\
\bottomrule
\end{tabular}
\end{center}
GSM8K is prompted in chat style with ``Let us think step by step'' and the final answer requested after the delimiter $\#\#\#\#$; reward extraction parses the final number. ARC-C uses a multiple-choice prompt listing choices A--D; reward checks the chosen letter. WinoGrande uses a sentence-completion prompt with two options A/B; reward checks the chosen option.

\subsection{Evaluation pool}
For each task, we construct a fixed evaluation pool of $P=320$ prompts by a deterministic random shuffle with \texttt{pool\_seed=0}. The same pool is used for all model sizes within a task to ensure comparability across scales. All main-text best-of-$N$ and best-of-30 point estimates are computed on this fixed pool (see below), with uncertainty quantified separately.

\subsection{Perturbations and candidate generation}
For parameters $\theta$, each ES probe candidate applies a single isotropic Gaussian perturbation
\begin{equation}
\theta' = \theta + \sigma\,u, \qquad u \sim \mathcal{N}(0, I),
\end{equation}
where $u$ matches the shapes of all model parameters. We do not normalize $u$ (so $\|u\|\approx\sqrt{d}$ in $d$ parameters). Perturbations are applied in-place and then exactly restored after evaluation; this is critical for numerical fidelity in \texttt{bfloat16}. We monitor restoration fidelity via baseline drift checks (below).

\paragraph{Perturbation scales.}
Main figures use $\sigma \in \{3\times 10^{-4}, 10^{-3}, 3\times 10^{-3}\}$; extended analyses additionally include $\sigma\in\{10^{-4},10^{-2}\}$.

\paragraph{Independent perturbation batches (for uncertainty).}
To avoid inflating extreme-value estimates via evaluation-set noise, we separate two sources of randomness:
(i) perturbation randomness (the ES sampling process) and (ii) evaluation-set randomness (finite prompt pool).
For each (task, model size, $\sigma$) condition, we generate $S$ independent perturbation batches, each containing $M=240$ candidates.
Candidate perturbation seeds are generated from a base seed (1234) using \texttt{rng.integers(0,2\^{}31,size=240)}; different batches use independent RNG streams.
In the main plots, uncertainty bars for best-of-$N$ are computed across these independent perturbation batches (Section~\ref{app:bestofN_estimation}).

\subsection{Reward computation and stored data}
For each candidate and prompt, we compute a binary reward $r_i\in\{0,1\}$. The mean reward on the pool is
\begin{equation}
R = \frac{1}{P}\sum_{i=1}^P r_i,
\qquad
\Delta R = R_{\text{candidate}} - R_{\text{baseline}}.
\end{equation}
We store per-candidate data as (i) the scalar $\Delta R$ on the full pool and (ii) packed bitstrings for the baseline and candidate per-prompt rewards (320 bits) encoded in base64. This enables exact reconstruction and re-aggregation under alternative evaluation procedures (e.g., prompt bootstrap for sensitivity analyses).

\subsection{Estimating best-of-$N$ without evaluation-noise inflation}
\label{app:bestofN_estimation}

Our main paper uses extreme-value statistics because selection is central to ES and because improvement directions are rare.
However, best-of-$N$ is sensitive to additional measurement noise: injecting evaluation-set resampling inside the maximization can inflate the maximum (a ``winner's curse'').
We therefore compute best-of-$N$ point estimates on the \emph{full fixed pool} and use independent perturbation batches to quantify uncertainty.

\paragraph{Best-of-$N$ point estimate (full pool).}
For each perturbation batch $s\in\{1,\dots,S\}$, we compute candidate deltas $\{\Delta R^{(s)}_j\}_{j=1}^M$ on the full pool. For a given population size $N$, we estimate
\begin{equation}
\Delta_N^{\ast,(s)}(\theta,\sigma)\;=\;\mathbb{E}\!\left[\max_{j\in \mathcal{S}} \Delta R^{(s)}_j\right],
\end{equation}
where $\mathcal{S}$ is a uniformly random subset of $\{1,\dots,M\}$ of size $N$ drawn \emph{without replacement}. We approximate this expectation by Monte Carlo subset sampling (typically 2000 subsets per $N$), which is computationally cheap since it operates only on stored scalar deltas.
We report the mean across batches, $\widehat{\Delta_N^\ast} = \frac{1}{S}\sum_s \Delta_N^{\ast,(s)}$.

\paragraph{Uncertainty (perturbation-batch variability).}
Error bars in the main-text best-of-$N$ curves are computed across the $S$ independent perturbation batches:
we report $\pm 1.96\times \mathrm{SE}$, where $\mathrm{SE}$ is the standard error of $\{\Delta_N^{\ast,(s)}\}_{s=1}^S$.
These uncertainty estimates reflect algorithmic variability due to perturbation sampling, which is the relevant uncertainty for population-requirement claims.

\paragraph{Population sizes.}
For best-of-$N$ curves we evaluate $N\in\{5,10,20,30,50\}$, with $N=30$ emphasized in the main text.

\paragraph{Evaluation-set uncertainty and extreme-value effects (appendix only).}
We additionally quantify evaluation-set uncertainty via bootstrap over prompts (resampling the $P=320$ prompts with replacement) \emph{without} changing the candidate set. This provides confidence intervals for statistics on the fixed pool.
We do \emph{not} use prompt bootstrap to define main-text best-of-$N$ point estimates because doing so introduces extra noise inside the maximization and can inflate maxima, consistent with classical extreme-value sensitivity (often modeled by Gumbel-type behavior for maxima).
We use prompt bootstrap and subset-size sensitivity studies only as robustness checks in the appendix.

\subsection{Other summary statistics}
\begin{figure}[t]
  \centering
  \includegraphics[width=0.8\columnwidth]{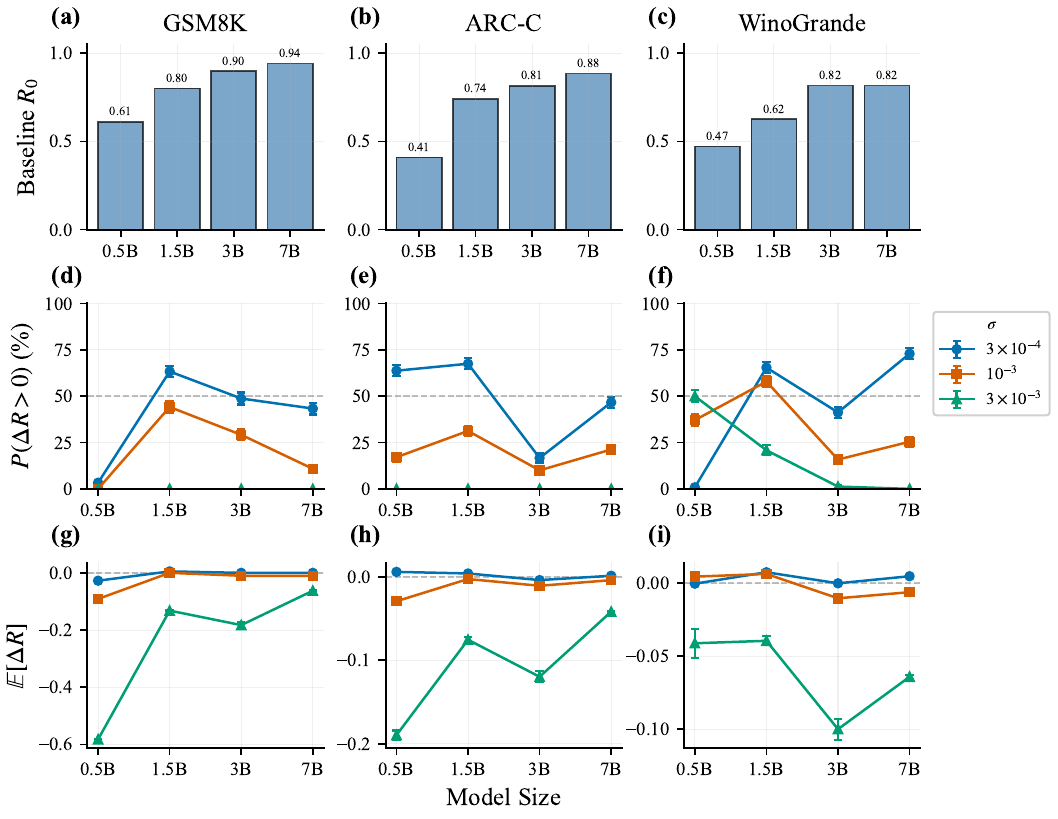}
  \caption{\textbf{Baseline performance and average perturbation statistics.}
  Columns correspond to GSM8K, ARC-C, and WinoGrande; rows show complementary summary statistics computed on the fixed evaluation pool ($P=320$ prompts).
  \textbf{Top row (a--c):} baseline accuracies $R_0$ for each model size.
  \textbf{Middle row (d--f):} probability of improvement $p(\Delta R>0)$ across perturbation scales $\sigma\in\{3\times10^{-4},10^{-3},3\times10^{-3}\}$.
  \textbf{Bottom row (g--i):} mean improvement $\mathbb{E}[\Delta R]$ across perturbations at the same $\sigma$ values.
  In many regimes the mean improvement is near zero or negative and $p(\Delta R>0)<\tfrac{1}{2}$, indicating that most perturbations are non-improving; this is expected in nonconvex, high-dimensional objectives and motivates extreme-value statistics (best-of-$N$) in the main text, which capture rare-but-meaningful improvements.}
  \label{fig:app_abc}
\end{figure}

\begin{figure}[t]
  \centering
  \includegraphics[width=0.8\columnwidth]{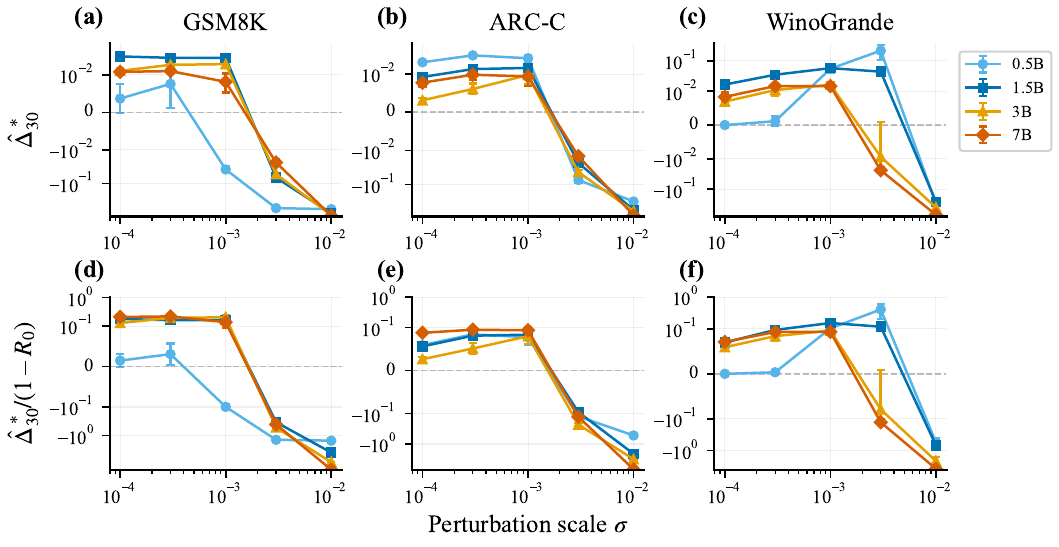}
  \caption{\textbf{Perturbation-scale sensitivity of best-of-30 improvements.}
  Best-of-30 expected improvement $\widehat{\Delta}^\ast_{30}$ (top row, a--c) and headroom-normalized best-of-30 $\widehat{\Delta}^\ast_{30}/(1-R_0)$ (bottom row, d--f) as a function of perturbation scale $\sigma$ for GSM8K, ARC-C, and WinoGrande.
  Moderate $\sigma$ values yield positive best-of-30 improvements across model sizes, while very large $\sigma$ can produce negative improvements, consistent with a scale-mismatch/variance-dominated regime.
  Error bars reflect uncertainty from independent perturbation batches (main text protocol), and headroom normalization controls for differing baseline saturation across model sizes.}
  \label{fig:app_de}
\end{figure}

We also report (i) $p(\mathrm{improve})=\Pr(\Delta R>0)$ and (ii) $\mathbb{E}[\Delta R]$ across candidates. Unless stated otherwise, these are computed on the full pool with uncertainty obtained by bootstrap over candidates (1000 replicates, resampling candidates with replacement).
Figure~\ref{fig:app_abc} summarizes baseline performance and average perturbation statistics ($p(\Delta R>0)$ and $\mathbb{E}[\Delta R]$), highlighting that most perturbations are non-improving even when best-of-$N$ is positive.
Figure~\ref{fig:app_de} shows sensitivity of best-of-30 to perturbation scale $\sigma$ (raw and headroom-normalized), illustrating the existence of an intermediate $\sigma$ regime where improvements are accessible.

\subsection{Headroom normalization}
To control for differences in baseline accuracy across model sizes, we report headroom-normalized improvements
\begin{equation}
\Delta^{\ast,\mathrm{rel}} = \frac{\Delta^\ast}{1-R_0},
\end{equation}
where $R_0$ is the baseline accuracy on the fixed pool. This measures improvement as a fraction of remaining achievable reward (given the binary accuracy ceiling at 1). The baseline accuracies for different model sizes and different tasks are shown in Figure~\ref{fig:app_abc}.

\paragraph{Headroom-normalized improvement distributions.}
To complement the best-of-$N$ summaries in the main text, we visualize the full distribution of perturbation outcomes.
For each task, model size, and perturbation scale $\sigma$, we sample $M=240$ random weight perturbations and compute the headroom-normalized change
$\Delta R/(1-R_0)$, where $\Delta R = R(\theta+\sigma\varepsilon)-R_0$ and $R_0$ is the baseline accuracy on the fixed prompt pool.
Headroom normalization enables fair comparison across models with different baseline accuracies: a value of $0.1$ corresponds to capturing $10\%$ of the remaining possible improvement under a binary reward ceiling.
Across tasks, we observe a ``locality window'' in $\sigma$: sufficiently small perturbations can retain a nontrivial improving tail, while larger $\sigma$ rapidly shifts the distribution negative and drives $p(\Delta R>0)$ toward zero.

\begin{figure*}[t]
  \centering
  \includegraphics[width=0.98\textwidth]{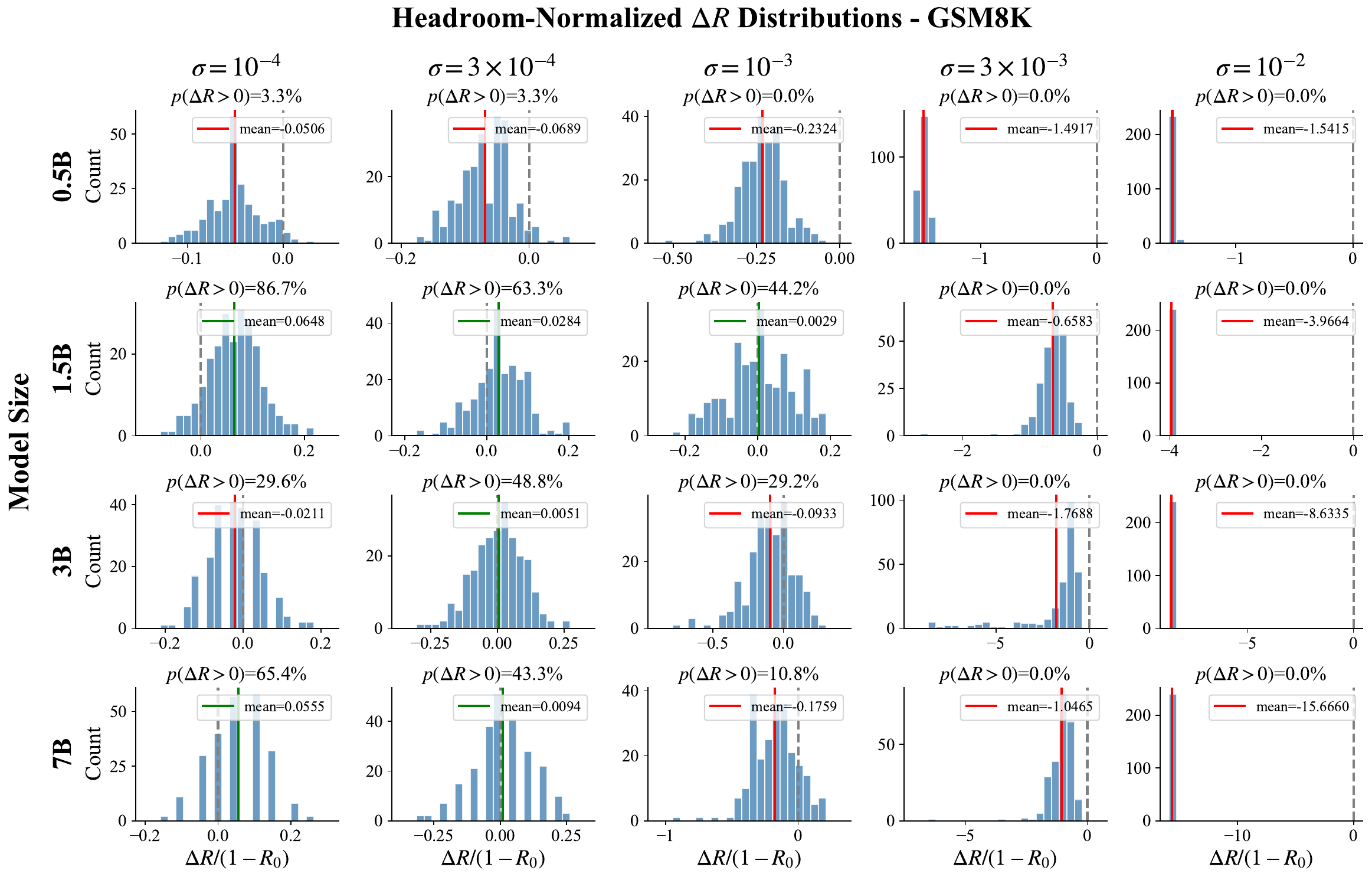}
  \caption{\textbf{Headroom-normalized perturbation outcome distributions on GSM8K.}
  Each panel shows $\Delta R/(1-R_0)$ for $M=240$ random perturbations, with the same conventions as Fig.~\ref{fig:headroom_deltaR_arcc}.
  Across model sizes, sufficiently small $\sigma$ can preserve a nontrivial improving tail, while increasing $\sigma$ shifts the distribution left and drives $p(\Delta R>0)$ toward zero, consistent with leaving the local regime where improvements are accessible.}
  \label{fig:headroom_deltaR_gsm8k}
\end{figure*}

\begin{figure*}[t]
  \centering
  \includegraphics[width=0.98\textwidth]{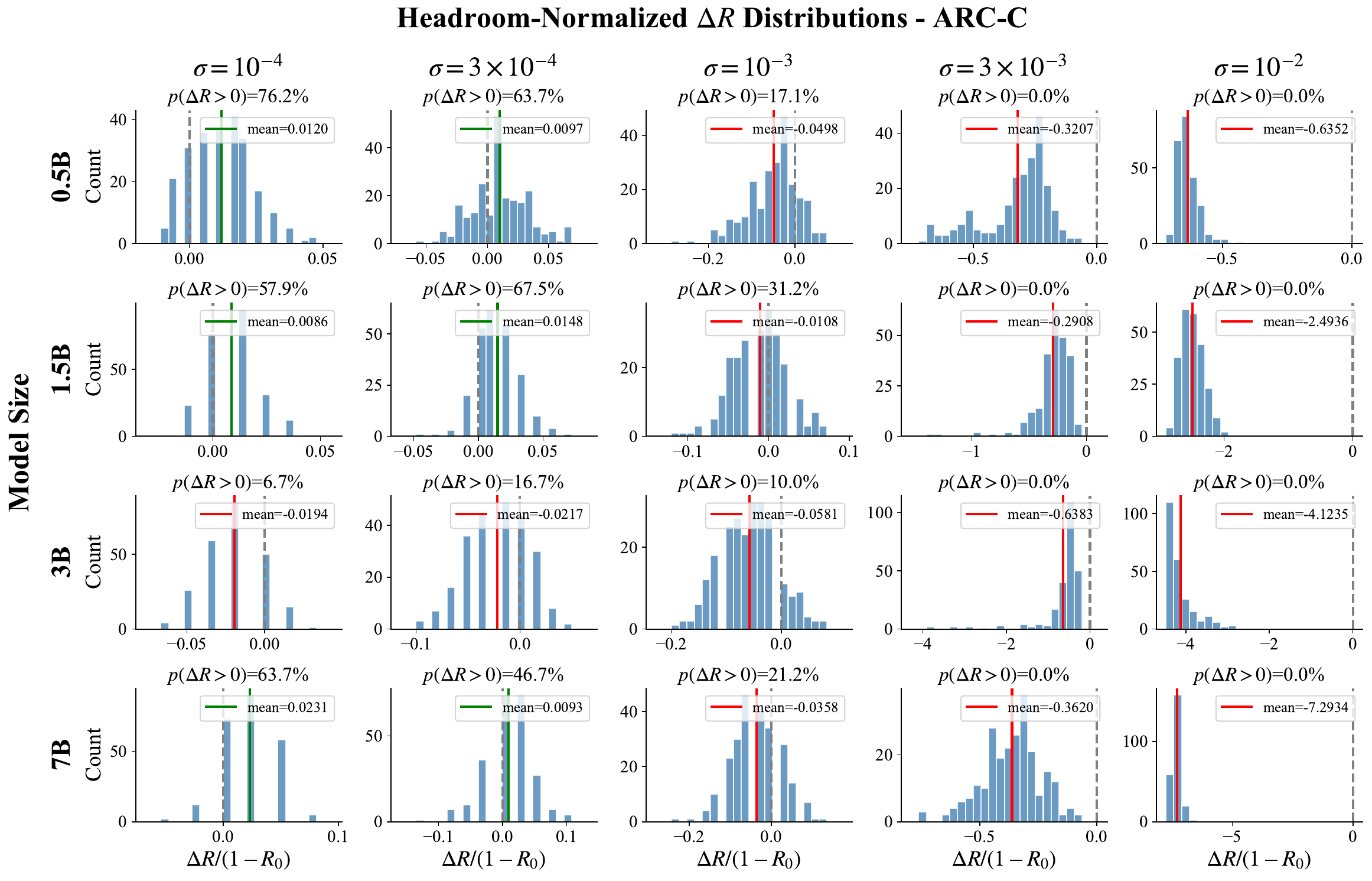}
  \caption{\textbf{Headroom-normalized perturbation outcome distributions on ARC-C.}
  Each panel shows the distribution of $\Delta R/(1-R_0)$ over $M=240$ random weight perturbations, where $\Delta R=R_i-R_0$ is the change in accuracy from baseline $R_0$ and $(1-R_0)$ is the remaining headroom.
  Rows correspond to model sizes (0.5B--7B) and columns to perturbation scale $\sigma$.
  The gray dashed vertical line marks $\Delta R=0$ (no change); the solid vertical line marks the mean (green if positive, red if negative).
  Panel titles report $p(\Delta R>0)$, the fraction of perturbations that improve over baseline.
  Headroom normalization makes the distributions comparable across model sizes and highlights a viable small-$\sigma$ regime where an improving tail exists versus larger-$\sigma$ regimes where perturbations predominantly degrade performance.}
  \label{fig:headroom_deltaR_arcc}
\end{figure*}

\begin{figure*}[t]
  \centering
  \includegraphics[width=0.98\textwidth]{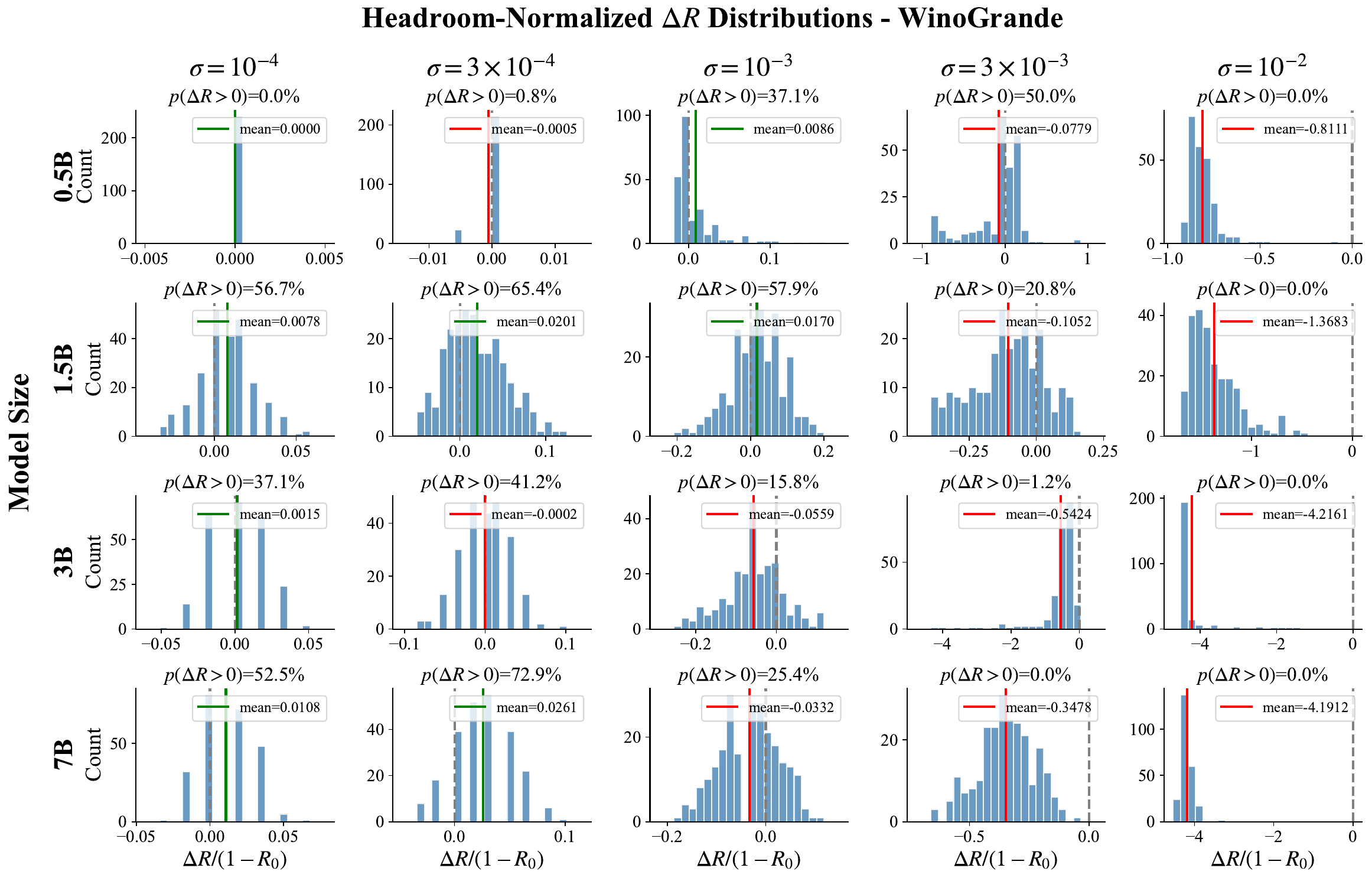}
  \caption{\textbf{Headroom-normalized perturbation outcome distributions on WinoGrande.}
  Each panel shows $\Delta R/(1-R_0)$ for $M=240$ random perturbations, with the same conventions as Fig.~\ref{fig:headroom_deltaR_arcc}.
  The distributions emphasize that best-of-$N$ improvements arise from the right tail even when the mean is near zero or negative, and that this tail is strongly $\sigma$-dependent.}
  \label{fig:headroom_deltaR_winogrande}
\end{figure*}

\subsection{Auxiliary diagnostics: saturation population and tail quantiles}
\label{app:blessing_diag}

\begin{figure*}[t]
  \centering
  \includegraphics[width=0.8\textwidth]{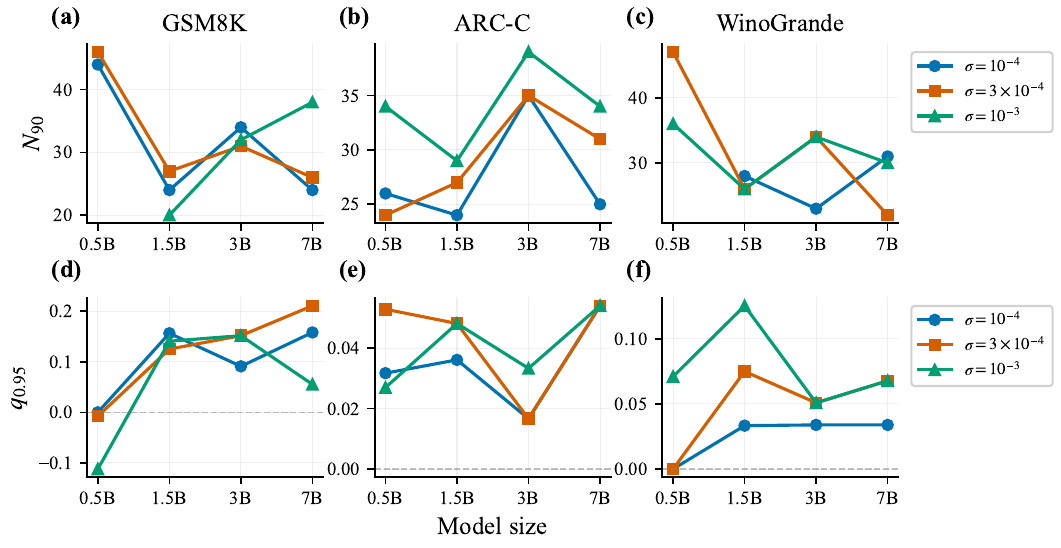}
  \caption{\textbf{Auxiliary accessibility diagnostics across model scale.}
  \textbf{Top row (a--c):} Saturation population $N_{90}$ as a function of model size for GSM8K, ARC-C, and WinoGrande at three perturbation scales $\sigma$.
  Here $N_{90}$ is the smallest population size $N$ such that the expected best-of-$N$ improvement reaches $90\%$ of the value at the largest evaluated population, i.e.,
  $N_{90}\defeq \min\{N:\Delta_N^*(\sigma)\ge 0.9\,\Delta_{N_{\max}}^*(\sigma)\}$.
  \textbf{Bottom row (d--f):} Upper-tail quantile $q_{0.95}$ of the headroom-normalized improvement distribution $\Delta R/(1-R_0)$ for the same conditions, estimated from the candidate pool of $M=240$ perturbations.
  (The dashed gray line marks $0$.)
  While these summaries are task- and $\sigma$-dependent and not monotonic in model size, they show no systematic increase in the population required to access the improving tail as model size grows from 0.5B to 7B.
  This provides an additional check, complementary to Fig.~\ref{fig:es_scaling_combined}, that improvement accessibility does not deteriorate in proportion to ambient parameter count.}
  \label{fig:app_accessibility_diag}
\end{figure*}

To complement the best-of-$N$ curves in the main text, we summarize improvement accessibility with two scalar diagnostics.
First, we define the \emph{saturation population} $N_{90}$ as the smallest population size needed to obtain $90\%$ of the expected best-of-$N$ improvement at the largest evaluated population, $N_{\max}$.
Second, we report an upper-tail statistic of the perturbation-induced improvement distribution, the $95$th percentile $q_{0.95}$ of $\Delta R/(1-R_0)$ across the candidate pool.
Figure~\ref{fig:app_accessibility_diag} shows that both $N_{90}$ and $q_{0.95}$ vary across tasks and perturbation scales, but neither exhibits a systematic increase with model size.
These diagnostics provide an additional, compact check that the population required to access improving perturbations does not grow proportionally with the number of model parameters.

\subsection{Reward-based curvature evidence via perturbation SLQ}
\label{app:reward_curvature_slq}

Our main-text blessing-of-dimensionality hypothesis is formulated in \emph{reward geometry} terms: the number of curvature-active directions governing local improvement need not grow proportionally with parameter count.
To complement the operational best-of-$N$ accessibility probes, we directly estimate curvature structure of the \emph{reward-defined} objective at a fixed smoothing scale by analyzing the Hessian of $\J_\sigma$.

\begin{figure}[t]
  \centering
  \includegraphics[width=0.6\columnwidth]{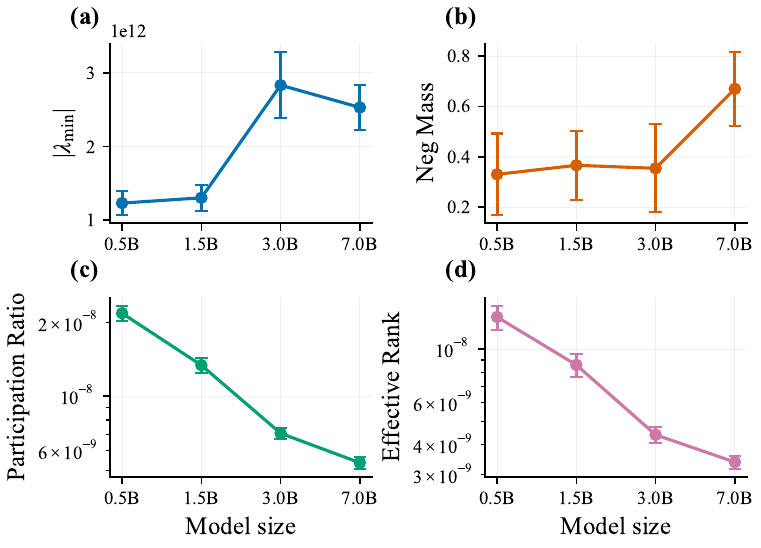}
  \caption{\textbf{Reward-based curvature proxy across model scale (GSM8K): concentration of curvature-active structure.}
  We estimate Hessian-spectrum-derived metrics of the \emph{ES-smoothed reward objective}
  $\J_\sigma(\theta)=\E_{\varepsilon\sim\mathcal N(0,I)}[\J(\theta+\sigma\varepsilon)]$
  using a perturbation-based Hessian--vector product estimator (no autograd) and stochastic Lanczos quadrature (SLQ).
  Rewards are binary GSM8K training accuracy on a fixed subset of 100 prompts with a fixed prompting and answer-extraction rule.
  \textbf{(a)} magnitude of the most negative eigenvalue $|\lambda_{\min}|$,
  \textbf{(b)} negative spectral mass (total weight on $\lambda<0$),
  \textbf{(c)} participation ratio, and
  \textbf{(d)} effective rank (both concentration measures; smaller values indicate fewer directions carry most spectral weight).
  Across Qwen2.5-Instruct model sizes (0.5B--7B), participation ratio and effective rank decrease, indicating that curvature-relevant structure in the reward-defined landscape becomes increasingly \emph{concentrated} rather than expanding in proportion to parameter count.
  Error bars show mean $\pm$ seed variability over 5 random seeds.}
  \label{fig:reward_slq_metrics_gsm8k}
\end{figure}

Figure~\ref{fig:reward_slq_metrics_gsm8k} reports Hessian-spectrum-derived summaries of $\nabla^2 \J_\sigma(\theta)$ for GSM8K across model sizes.
The key observation is that concentration measures (participation ratio and effective rank) decrease with model size, indicating that a smaller set of directions carries most curvature-relevant structure in the ES-smoothed reward landscape as models scale.
At the same time, the magnitude of extreme negative curvature can increase.
Together, these findings support a curvature-based blessing-of-dimensionality interpretation: scaling primarily adds weakly curved directions while the curvature-active structure remains relatively concentrated, consistent with the ``bulk + tail'' picture used in the main text.

\end{document}